\documentclass[journal]{IEEEtran}
\usepackage{graphicx}
\usepackage{amsmath, amssymb}
\usepackage{caption}
\usepackage{subcaption}
\usepackage{multirow}
\usepackage{soul,color}
\usepackage{cite}
\usepackage{pgfplots}
\pgfplotsset{width=10cm,compat=1.9}
\begin{document}
\title{Transformers Fusion across Disjoint Samples for Hyperspectral Image Classification}
\author{Muhammad Ahmad, Manuel Mazzara, Salvatore Distifano
	\thanks{M. Ahmad is with the Department of Computer Science, National University of Computer and Emerging Sciences, Islamabad, Chiniot-Faisalabad Campus, Chiniot 35400, Pakistan, and Dipartimento di Matematica e Informatica---MIFT, University of Messina, Messina 98121, Italy; (e-mail: mahmad00@gmail.com)}
	\thanks{M. Mazzara is with the Institute of Software Development and Engineering, Innopolis University, Innopolis, 420500, Russia. (e-mail: m.mazzara@innopolis.ru)}
	\thanks{S. Distefano is with  Dipartimento di Matematica e Informatica---MIFT, University of Messina, Messina 98121, Italy. (e-mail: sdistefano@unime.it)}
}
\markboth{Journal of \LaTeX\ Class Files,~Vol.~00, No.~00, ~2024}
{M. Ahmad \MakeLowercase{\textit{et al.}}:}
\maketitle
\begin{abstract}
3D Swin Transformer (3D-ST) known for its hierarchical attention and window-based processing, excels in capturing intricate spatial relationships within images. Spatial-spectral Transformer (SST), meanwhile, specializes in modeling long-range dependencies through self-attention mechanisms. Therefore, this paper introduces a novel method: an attentional fusion of these two transformers to significantly enhance the classification performance of Hyperspectral Images (HSIs). What sets this approach apart is its emphasis on the integration of attentional mechanisms from both architectures. This integration not only refines the modeling of spatial and spectral information but also contributes to achieving more precise and accurate classification results. The experimentation and evaluation of benchmark HSI datasets underscore the importance of employing disjoint training, validation, and test samples. The results demonstrate the effectiveness of the fusion approach, showcasing its superiority over traditional methods and individual transformers. Incorporating disjoint samples enhances the robustness and reliability of the proposed methodology, emphasizing its potential for advancing hyperspectral image classification.
\end{abstract}
\IEEEpeerreviewmaketitle
\section{Introduction}
\label{Intr}

\IEEEPARstart{H}{yperspectral Image Classification (HSIC)} plays a pivotal role in various domains including remote sensing \cite{ahmad2021hyperspectral}, earth observation \cite{lodhi2018hyperspectral}, urban planning \cite{li2024HD}, agriculture \cite{lu2020recent}, forestry \cite{adao2017hyperspectral}, target/object detection \cite{li2023lrr}, mineral exploration \cite{bedini2017use}, environmental monitoring \cite{weber2018hyperspectral, stuart2019hyperspectral}, climate change \cite{pande2023application}, food processing \cite{khan2021hyperspectral, khan2020hyperspectral}, bakery products \cite{saleem2020prediction}, bloodstain identification \cite{butt2022fast, zulfiqar2021hyperspectral}, and meat processing \cite{ayaz2020hyperspectral, ayaz2020myoglobin}. The wealth of spectral information in HSIs, spanning numerous narrow bands, poses both challenges and opportunities for effective classification \cite{hong2024spectralgpt}. Recent years have witnessed the success of Neural Networks \cite{ahmad2020fast, 10423094, 9170817, 10409250, 10433668} in various computer vision tasks, there is a noticeable shift towards exploring the potential of Transformer models for advancing HSI analysis. 

The Swin Transformer (ST) \cite{yao2023extended, 10399798, 9868046, ZU2024108041} exhibits notable strengths, primarily rooted in its hierarchical attention mechanism. This mechanism empowers the model to capture information across different scales, efficiently analyzing both local and global features \cite{rs15153721}. Additionally, the adoption of windowing-based processing enhances the scalability of STs enabling the effective handling of large images with reduced computational complexity \cite{FAROOQUE2023107070, rs15143491}. In the realm of HSIC, STs have proven their mettle, showcasing state-of-the-art performance \cite{Selen2105668}. They have surpassed CNNs in specific scenarios, demonstrating their efficacy for HSIC tasks \cite{rs15102696}. However, despite these successes, STs are not without their limitations. While excelling in capturing spatial relationships, they may encounter challenges in dealing with sequential data, making tasks reliant on spectral dependencies less optimal for them \cite{10123084}. Furthermore, the hierarchical attention mechanism, while potent, introduces additional complexity. Training large ST models demands substantial computational resources, posing challenges for researchers with limited access to high-performance computing. Additionally, akin to other deep models, the interpretability of STs raises concerns, particularly in complex tasks like HSIC. Understanding the decision-making processes of these models remains an ongoing focus of research.

Similarly, the vision and spatial-spectral transformers (SSTs) \cite{10423821, rs13030498, 10409287, rs16020404, 10432978, 10422823, 10399888, 10387229} excel in capturing global contextual information. The self-attention mechanism allows the model to consider relationships between all HSI regions simultaneously, providing a holistic understanding of the visual context \cite{9895238}. Moreover, unlike CNNs, SSTs exhibit strong scalability to high-resolution HISs. They can effectively process large HSI datasets without the need for complex pooling operations. Thus, SSTs are versatile and have demonstrated success for HSIC and their architecture's adaptability contributes to their widespread applicability. Furthermore, SSTs alleviate the reliance on handcrafted features by learning hierarchical representations directly from raw pixel values \cite{9627165}. This end-to-end learning approach simplifies the model-building process and often leads to improved performance. Moreover, the attention maps generated by SSTs offer insights into the model's decision-making process. This interpretability is valuable for understanding which parts of the image contribute most to the specific predictions \cite{10387571}.

Despite the achievements of SSTs, they exhibit certain limitations. Notably, training large SSTs can be computationally demanding, particularly as the model size expands \cite{10419133, 10379170}. The self-attention mechanism introduces quadratic complexity concerning sequence length, potentially hindering scalability \cite{HUANG2024109897}. Unlike CNNs, which inherently possess translation invariance through shared weight convolutional filters, SSTs may encounter difficulties in capturing spatial relationships that remain invariant to small translations in the input \cite{SUN2024102163}. Moreover, the reliance of SSTs on dividing input images into fixed-size patches during the tokenization process might not efficiently capture fine-grained details \cite{10418237, 10400415}. The quadratic scaling of self-attention raises challenges, especially when dealing with long sequences. Additionally, optimal performance for SSTs often necessitates substantial amounts of training data. Attempting to train these models on smaller datasets might lead to overfitting, thereby limiting their effectiveness in scenarios with restricted labeled data \cite{ahmad2024importance, 10399798, ahmad2024pyramid, BUTT2024103773, ahmad2024traditional}.

In light of these limitations, potential solutions proposed in the literature include the integration of SSTs with other architectures. For example, a hybrid model has the potential to capitalize on the strengths of both approaches, merging the global context understanding of SSTs with the local feature extraction capabilities of CNNs \cite{10415455, 10400402}. However, one potential shortcoming of hybrid transformers is the added complexity in model architecture. Integrating different transformer variants or combining transformers with other architectures may result in increased intricacy, making the model harder to interpret and potentially requiring more computational resources for training and inference.

Additionally, there is room for exploration in optimizing the attention mechanism, with efforts focused on sparse attention patterns or alternative attention mechanisms that could address the challenges posed by computational complexity \cite{SHU2024107351, MA2024102148, rs16020404} capturing long-range dependencies and integrating global and local features in transformers is the challenge of striking a balance between model complexity and computational efficiency. While optimizing attention patterns can enhance performance, it may also increase computational demands, making the model more resource-intensive and potentially limiting its applicability in scenarios with constrained resources.

Extending the capabilities of SSTs to handle multi-modal inputs stands out as another promising research direction \cite{10379015}. This expansion could broaden their applicability and enhance their capacity to comprehend intricate relationships within diverse data types \cite{10382626, Miao22970, QU2024121363}. However, integrating information from different modalities requires careful consideration of feature representations and alignment. The model may face challenges in effectively learning meaningful cross-modal relationships, and designing architectures that can efficiently fuse and process diverse data sources remains an ongoing research challenge. Additionally, collecting and annotating large-scale multi-modal datasets for training can be resource-intensive. Lastly, the development of efficient tokenization strategies is crucial, aiming to capture fine-grained details while mitigating the quadratic scaling issue. Such strategies could notably enhance the performance of SSTs in tasks requiring high spatial resolution. In short, the ST and SST are both transformer-based architectures designed for image classification tasks, but they differ in their approach to handling image data. Here are some of the major differences between ST and SST:

\textbf{Image Patch Processing:} ST employs a hierarchical structure where it initially divides the image into non-overlapping patches, but then hierarchically processes these patches using a window-based self-attention mechanism. This allows it to capture both local and global context efficiently. SST divides the input image into fixed-size non-overlapping patches, linearly embeds each patch, and then flattens the 2D spatial information into a 1D sequence to feed into the transformer. \textbf{Hierarchical Self-Attention:} ST introduces a shifted window-based self-attention mechanism. Instead of attending to all positions equally, it uses local self-attention windows that hierarchically slide across the sequence. This allows the model to capture both local and long-range dependencies effectively. Whereas, SST uses a single self-attention mechanism across the entire sequence of patches. \textbf{Positional Encoding:} ST uses shifted windows in self-attention and does not rely on explicit positional embeddings. The local windows implicitly capture positional relationships. Whereas SST uses positional embeddings to provide the model with information about the spatial arrangement of patches.

\begin{figure}[!hbt]
    \centering
	\begin{subfigure}{0.24\textwidth}
		\includegraphics[width=0.99\textwidth]{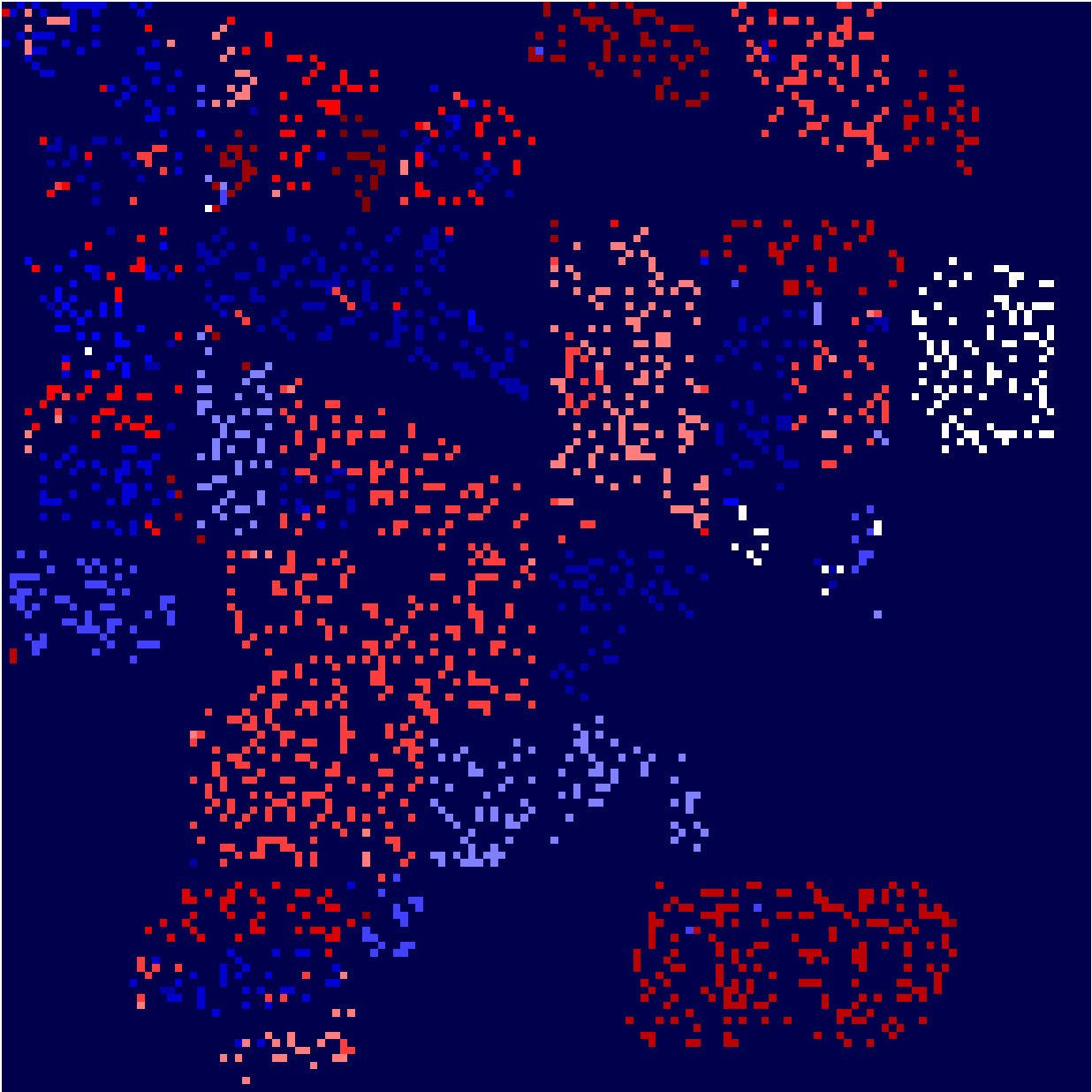}
		\caption{3D Swin Transformer} 
		\label{FigAA}
	\end{subfigure}
	\begin{subfigure}{0.24\textwidth}
		\includegraphics[width=0.99\textwidth]{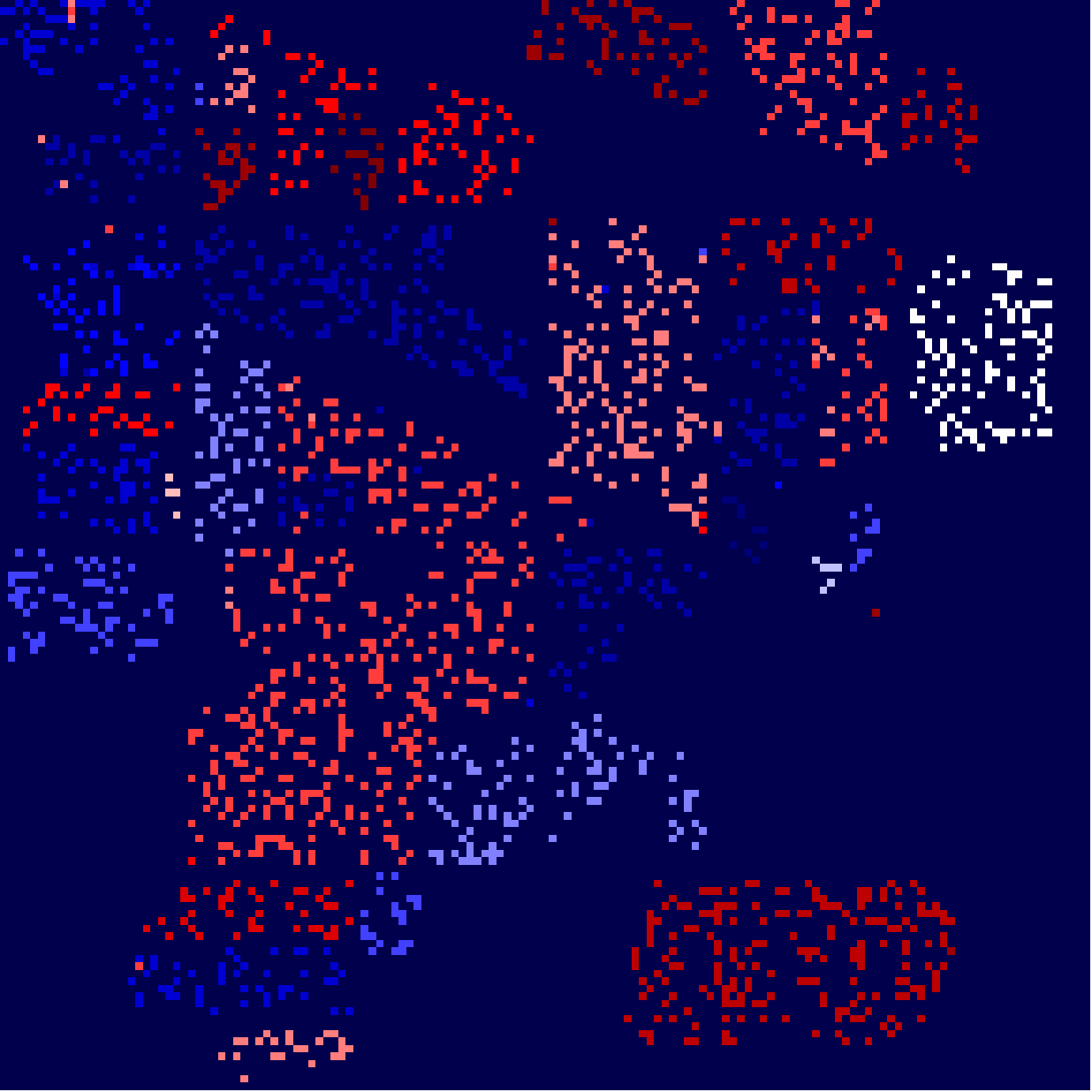}
		\caption{Spatial-Spectral Transformer}
		\label{FigBB}
	\end{subfigure} 
\caption{Land cover maps for the disjoint test set. Comprehensive class-wise results can be found in Table \ref{tab:my_label}.}
\label{FigA}
\end{figure}

Given the distinctions outlined above, we proceeded to conduct preliminary experimental evaluations focusing on individual transformer models using the Indian Pines dataset. Both models underwent training using an $8 \times 8$ patch size, with 70\% of the samples allocated for training, 20\% for validation, and the remaining 10\% reserved for testing purposes which, in general, exceed thresholds suitable for trivial model evaluation. Upon reviewing the results presented in the provided Table \ref{tab:my_label} and Figure \ref{FigA}, it is evident that the individual transformer models fail to achieve accuracy levels commensurate with expectations, even when trained on a substantially larger number of samples. Consequently, in light of these findings, this study proposes the fusion of these two transformers to achieve superior results while utilizing fewer training samples. This phenomenon is elaborated upon in the experimental results and discussion section.

\begin{table}[!hbt]
    \centering
    \caption{Experimental results were conducted on disjoint training, validation, and test sets using the Indian Pines dataset for both the 3D Swin Transformer (3D ST) and the Spatial-Spectral Transformer (SST).}
    \begin{tabular}{c|cc|cc} \hline 
        \multirow{2}{*}{\textbf{Class}} & \multicolumn{2}{c|}{3D Swin Transformer} & \multicolumn{2}{c}{SS Transformer} \\ \cline{2-5}
        & Val & Test & Val & Test \\ \hline 
        Alfalfa & --- & --- & 89.02 & 89.50 \\
        Corn notill & 84.8396 & 86.7132 & 96.2099 & 96.8531 \\
        Corn mintill & 77 & 75.3012 & 95.50 & 93.3734 \\
        Corn & 61.4035 & 66.6666 & 87.7192 & 89.5833 \\
        Grass pasture & 96.5517 & 92.7835 & 99.1379 & 98.9690 \\
        Grass trees & 98.2954 & 97.2602 & 100 & 99.3150 \\
        Grass mowed & --- & --- & 92.2861 & 93.0124 \\
        Hay windrowed & 100 & 100 & 100 & 100 \\
        Oats & --- & --- & 64.0460 & 65.1034 \\
        Soybean notill & 76.9230 & 80 & 95.2991 & 93.8461 \\
        Soybean mintill & 95.9322 & 94.7046 & 97.2881 & 97.5560 \\
        Soybean clean & 71.3286 & 70.5882 & 98.6013 & 100 \\
        Wheat & 100 & 97.5609 & 100 & 100 \\
        Woods & 96.3815 & 97.2332 & 99.6710 & 98.8142 \\
        Buildings & 82.7956 & 88.4615 & 100 & 100 \\
        Stone-Steel & 95.6521 & 100 & 100 & 100 \\ \hline 
        \textbf{Kappa} & 86.2182 & 86.4678 & 94.2280 & 95.0067 \\ 
        \textbf{OA} & 87.9610 & 88.1751 & 94.5678 & 94.3722 \\ \hline
    \end{tabular}
    \label{tab:my_label}
\end{table}

Therefore, this work proposed an attention-based feature fusion of 3D ST and SST which presents a powerful synergy for HSIC. This fusion enhances the model's contextual understanding by capturing long-range dependencies and integrating global and local features effectively. The combined model demonstrates adaptability, leveraging the strengths of both transformers. It excels in handling spatial and sequential relationships, reducing the reliance on handcrafted features, and offering interpretable decision-making through attention maps. With scalability to high-resolution images and demonstrated state-of-the-art performance, this fusion emerges as a versatile and impactful solution for a wide spectrum of computer vision applications. In short, the contributions of this work can be encapsulated as follows:

\begin{enumerate}
    \item \textbf{Synergistic Fusion of 3D ST and SST:} This paper proposes the fusion of 3D ST and SST, harnessing the complementary strengths of hierarchical attention, window-based processing, and long-range dependency modeling. The synergy between these two transformers is a novel contribution with the potential to advance HSIC. The fusion approach significantly contributes to achieving more precise and accurate classification results in HSIs. By leveraging the strengths of both transformers, our methodology enhances the discriminative power of the model, leading to superior classification performance.

    \item \textbf{Integrated Attention Mechanisms for Enhanced Modeling:} A distinctive feature of our approach is the seamless integration of attentional mechanisms from both 3D ST and SST. This integration refines the modeling of spatial and spectral information, elevating the capacity of the model to capture intricate details crucial for HSIC.

    \item \textbf{Importance of Disjoint Sample Utilization:} A notable contribution lies in emphasizing the importance of employing disjoint training, validation, and test samples. This methodological choice enhances the reliability of our experimental evaluations, ensuring a more rigorous assessment of the fusion approach's performance. The incorporation of disjoint samples not only enhances the robustness of our methodology but also contributes to its overall reliability. This emphasis on sample separation reinforces the generalizability of the proposed fusion approach, underlining its potential for advancing the field of HSIC.
\end{enumerate}

In a nutshell, ST excels in capturing intricate spatial relationships within images, emphasizing its strength in spatial modeling. On the other hand, SST specializes in modeling long-range dependencies through self-attention mechanisms, with a primary focus on spectral information. While SST is designed for handling spectral complexities through self-attention, ST is effective in capturing spatial relationships. The fusion of ST and SST offers a comprehensive solution, addressing challenges associated with both spatial and spectral complexities in Hyperspectral Images (HSIs). This integration enhances the recognition of patterns and structures, potentially leading to the development of more efficient tokenization strategies that capture fine-grained details in HSIs. Additionally, this fusion helps mitigate challenges related to tokenization and alleviates issues associated with the scaling of self-attention concerning sequence length.

\section{Proposed Methodology}
\label{PM}

\begin{figure*}[!hbt]
    \centering
    \includegraphics[width=0.98\textwidth]{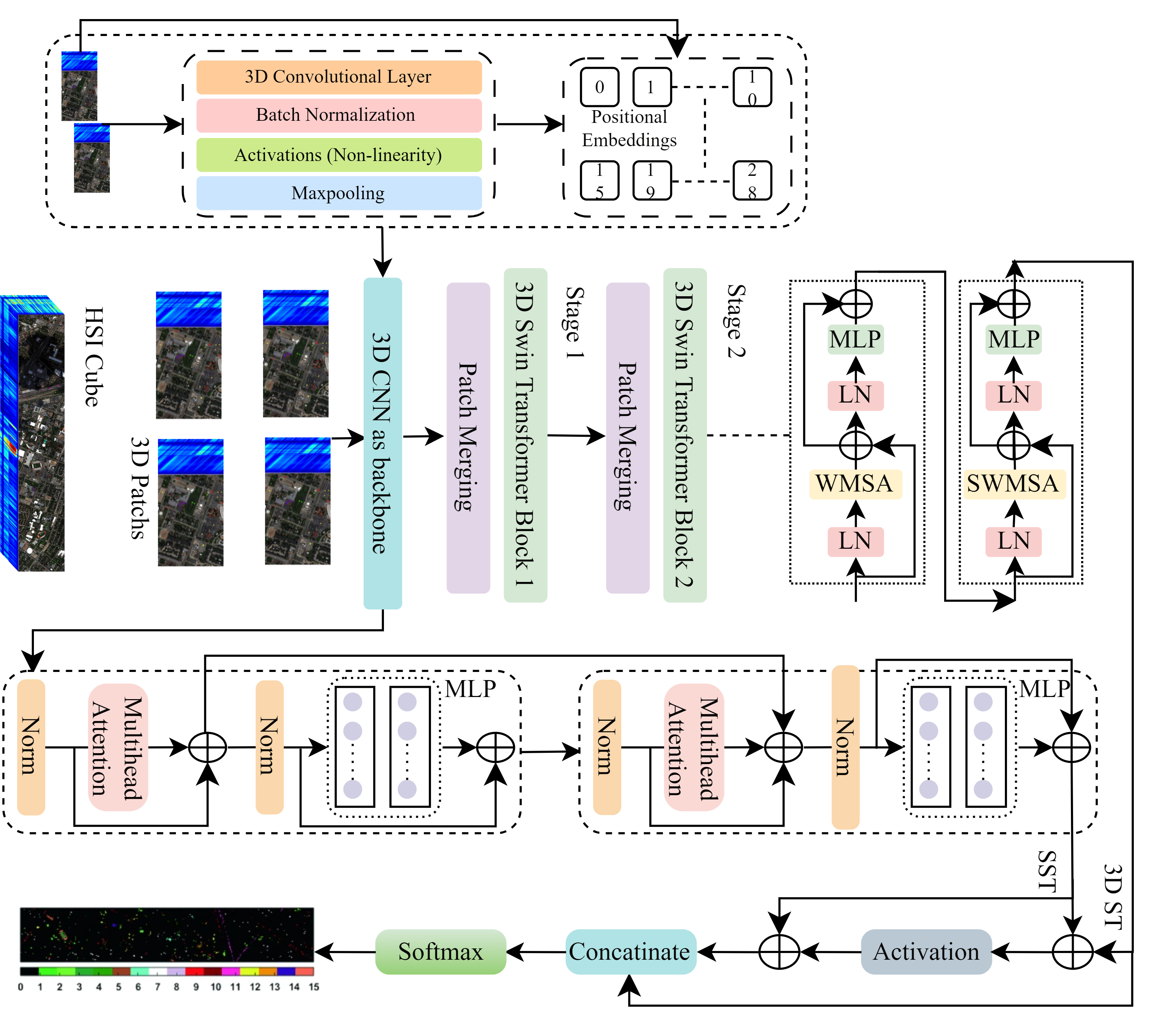}
    \caption{The comprehensive structure of 3D ST and SST. 3D ST comprises six stages, and 3D ST produces feature maps at varying scales, with each stage contributing to the output. Within 3D ST blocks, components include LN (Layer Normalization), 3D WMSA (3D Window-based Multi-Head Self-Attention), MLP (Multi-Layer Perceptron), and residual connections. In contrast, SST consists of four transformer layers and a feature fusion module.}
    \label{Fig1}
\end{figure*}

An HSI cube can be represented as $X = \{x_i, y_i\} \in \mathcal{R}^{(M \times N \times B)}$, where each $x_i = \{x_{i,1}, x_{i,2}, x_{i,3}, \dots, x_{i,L}\}$, and $y_i$ denotes the class label of each $x_i$. The HSI data cube $X$ undergoes an initial division into overlapping 3D patches. Each patch, centered at a spatial location $(\alpha, \beta)$, spans a spatial extent of $S \times S$ pixels across all $B$ bands. The total count of 3D patches ($m$) extracted from $X$ (i.e., $X \in \mathbb{R}^{(S \times S \times B)}$) is $(M-S+1) \times (N-S+1)$. A patch located at $(\alpha, \beta)$ is represented as $P_{\alpha, \beta}$ and covers spatial dimensions from $\alpha - \frac{S-1}{2}$ to $\alpha + \frac{S-1}{2}$ in width and $\beta - \frac{S-1}{2}$ to $\beta + \frac{S-1}{2}$ in height \cite{10399798}. The labels assigned to these patches are determined by the label assigned to the central pixel within each respective patch. Later these patches and the respective class labels were randomly allocated to $X_{train}$, as training data. The validation set, $X_{val}$, and test set $X_{test}$. Importantly, we ensure that $|X_{train}| \ll |X_{val}|$ and $|X_{train}| \ll |X_{test}|$, maintaining a significant distinction in sample sizes. Additionally, we guarantee that $X_{train} \cap X_{val} \cap X_{test} = \emptyset$, ensuring that training, validation, and test sets remain free of sample overlaps. This strict disjointedness is crucial to prevent biases and uphold the integrity of each dataset.

In transformer-based HSIC tasks, researchers frequently utilize convolutional layers to extract spatial-spectral semantic features from HSI patches. These features are subsequently mapped to tokens and input into the transformer encoder. Similarly, in this work, we initially spatial-spectral representation extraction is performed by a 3DCNN. Therefore, each patch with dimensions $WS \times WS \times B$ is processed using a 3DCNN comprising a convolutional layer (kernel size = ($B \times 1 \times WS \times WS$) with same pending and stride) with ReLu as the activation layer, batch normalization, and a max-pooling layer. The activation maps for the spatial-spectral position $(x, y, z)$ at the $i$-th feature map and $j$-th layer can be denoted as $v_{i,j}^{(x, y, z)}$ in which $d_{i - 1}$ represents the total number of feature maps at the $(i - 1)$-th layer, $w_{i, j}$ and $b_{i, j}$ denote the depth of the kernel and bias, respectively. Additionally, $2\gamma +1$, $2\delta +1$, and $2\nu + 1$ correspond to the height, width, and depth of the kernel \cite{ahmad2020fast, 9767615}.

\begin{multline}
v_{i,j}^{x,y,z} = ReLu \bigg(\sum_{\tau = 1}^{d_{i-1}} \sum_{\rho = -\gamma}^{\gamma} \sum_{\phi = -\delta}^{\delta} \sum_{\lambda = -\nu}^{\nu}  w_{i, j, \tau}^{\rho, \phi, \lambda} \times \\ v_{(i-1), \tau}^{(x+\rho), (y+\phi), (z+\lambda)} + b_{i,j} \bigg)
\end{multline}


The Swin Transformer (ST) excels in constructing multiscale feature maps by iteratively fusing neighboring patches using the window partition mechanism. Its linear computational complexity concerning image size proves advantageous for dense prediction tasks and high-resolution images. In this investigation, we used the ST architecture into a three-dimensional structure, denoted as 3D ST, tailored to accommodate the three-dimensional properties inherent in HSIs and effectively capture its rich spatial and spectral information. Figure \ref{Fig1} illustrates the architecture of the 3D ST with slight modifications to the original paper \cite{9868046}. Notably, compared to ST, the enhancements introduced are briefly summarized in the following aspects.

We characterize each HSI patch as $(WS \times WS \times B \times 1)$, where $WS$ denotes the patch size (window size), representing the height and width of the patch, with a feature dimension of 96. Subsequently, a positional and linear embedding layer projects these patches into arbitrary dimensions and a 3D CNN layer is deployed with ReLu as an activation function. During the patch merging phase, neighboring patches are combined while preserving the spectral dimensions. The key distinction between ST and 3D ST blocks lies in the window-based multi-head self-attention mechanism. 3D ST introduces the spectral domain to the multi-head self-attention, resulting in 3D multi-head self-attention. This incorporation considers the window partitioning and shifting mechanism as explained in \cite{9868046}. While ST employs 2D windows of size $(WS \times WS)$ to divide input patches evenly, 3D ST utilizes 3D windows sized $(WS \times WS \times P)$.

The 3D ST model utilized in this study comprises six stages, each comprising a patch merging module and a series of 3D ST blocks. As previously discussed, the patch merging module downsamples only the spatial dimension, retaining the spectral dimension, to concatenate neighboring patches into a larger patch. Simultaneously, a linear layer is employed to project the concatenated dimension to half of its original size. Subsequently, the 3D ST blocks extract self-attention information, maintaining the input resolution throughout this process. All other components within the 3D ST blocks remain consistent with the original ST, including components such as MLP, layer normalization, and residual connections as shown in Figure \ref{Fig1}. 

While the Spatial-Spectral Transformer (SST) shares the fundamental concept of treating an image as a sequence of non-overlapping patches, akin to tokens in Natural Language Processing (NLP), it employs a unique approach. Each image patch undergoes linear embedding, transforming it into a high-dimensional space and generating a sequence of feature vectors, serving as input tokens for the SST model. Let $X \in \mathbb{R}^{N \times D}$ represent the input tensor, where $N$ is the number of patches, and $D$ is the dimensionality of each patch after linear embedding.

To address the absence of inherent positional information in SST, positional encoding $PE \in \mathbb{R}^{N \times D}$ is used. This encoding is added to the input embeddings, enriching the model with spatial arrangement details. The core architecture of SST revolves around the transformer encoder, comprising multiple layers of self-attention mechanisms and a feedforward neural network (MLP). The self-attention mechanism plays a pivotal role in enabling the model to capture intricate relationships between distinct patches. Specifically, for a given input $H^{(0)} = X + PE$, each layer within the transformer encoder encompasses:

\begin{equation}
    H_{att}^{l} = SelfAttention(H^{(l-1)})
    \label{eq2}
\end{equation}

\begin{equation}
    H_{ff}^{l} = FeedForward(H_{att}^{l})
    \label{eq3}
\end{equation}

\begin{equation}
    H^{l} = H^{(l-1)} + H_{ff}^{l}
    \label{eq4}
\end{equation}
where equation \ref{eq2}, \ref{eq3}, and \ref{eq4} define the self-attention, FeedForward, and residual connections within the transformer layer. Let $H^{L'}$ (ST) and $H^{L}$ (SST) be the output tensors from ST and SST, respectively. First, the attention weights are computed element-wise by multiplying the corresponding elements of $H^{L}$ and $H^{L'}$ followed by an activation function as;

\begin{equation}
    Atten\_W = Activation(H^{L} \bigodot H^{L'})
\end{equation}

Later, the attention weights are applied element-wise to $H^{L'}$ as: 

\begin{equation}
    H^{L'}\_W = Atten\_W \bigodot H^{L'}
\end{equation}

The original features $H^{L}$ and the attended $H^{L'}$ are concatenated along the last axis followed by a softmax as: 

\begin{equation}
    Y = Softmax([H^{L}, H^{L'}\_W])
\end{equation}

The final tensor $Y$ represents the results of attentional feature fusion between $H^{L}$ and $H^{L'}$. 

\section{Experimental Results and Discussion}
\label{Exp}

The comprehensive comparative results are conducted against the state-of-the-art methods on widely utilized public HSI datasets to assess classification performance.

\subsection{HSI Data Description}

\begin{figure}[!hbt]
    \centering
	\begin{subfigure}{0.48\textwidth}
        \centering
		\includegraphics[width=0.99\textwidth]{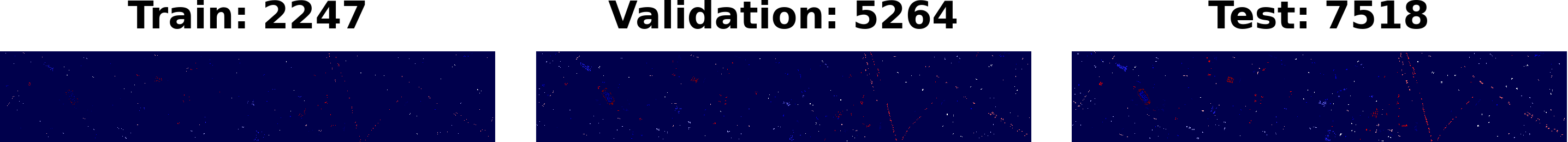}
		\caption{University of Houston (UH) Dataset} 
		\label{Fig2A}
	\end{subfigure}
    \begin{subfigure}{0.48\textwidth}
		\centering
        \includegraphics[width=0.99\textwidth]{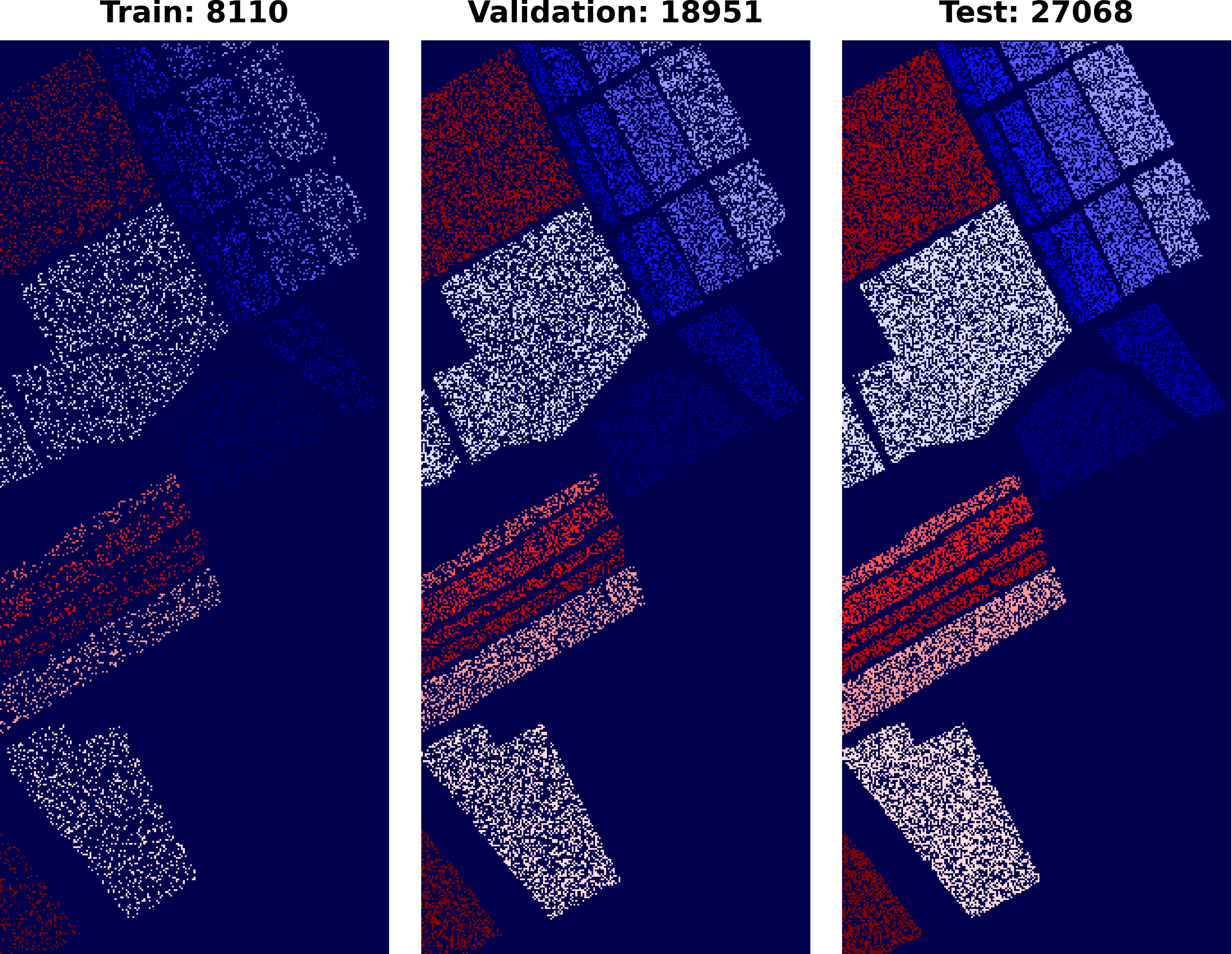}
		\caption{Salinas (SA) Dataset} 
		\label{Fig2C}
	\end{subfigure}
	\begin{subfigure}{0.48\textwidth}
		\centering
        \includegraphics[width=0.99\textwidth]{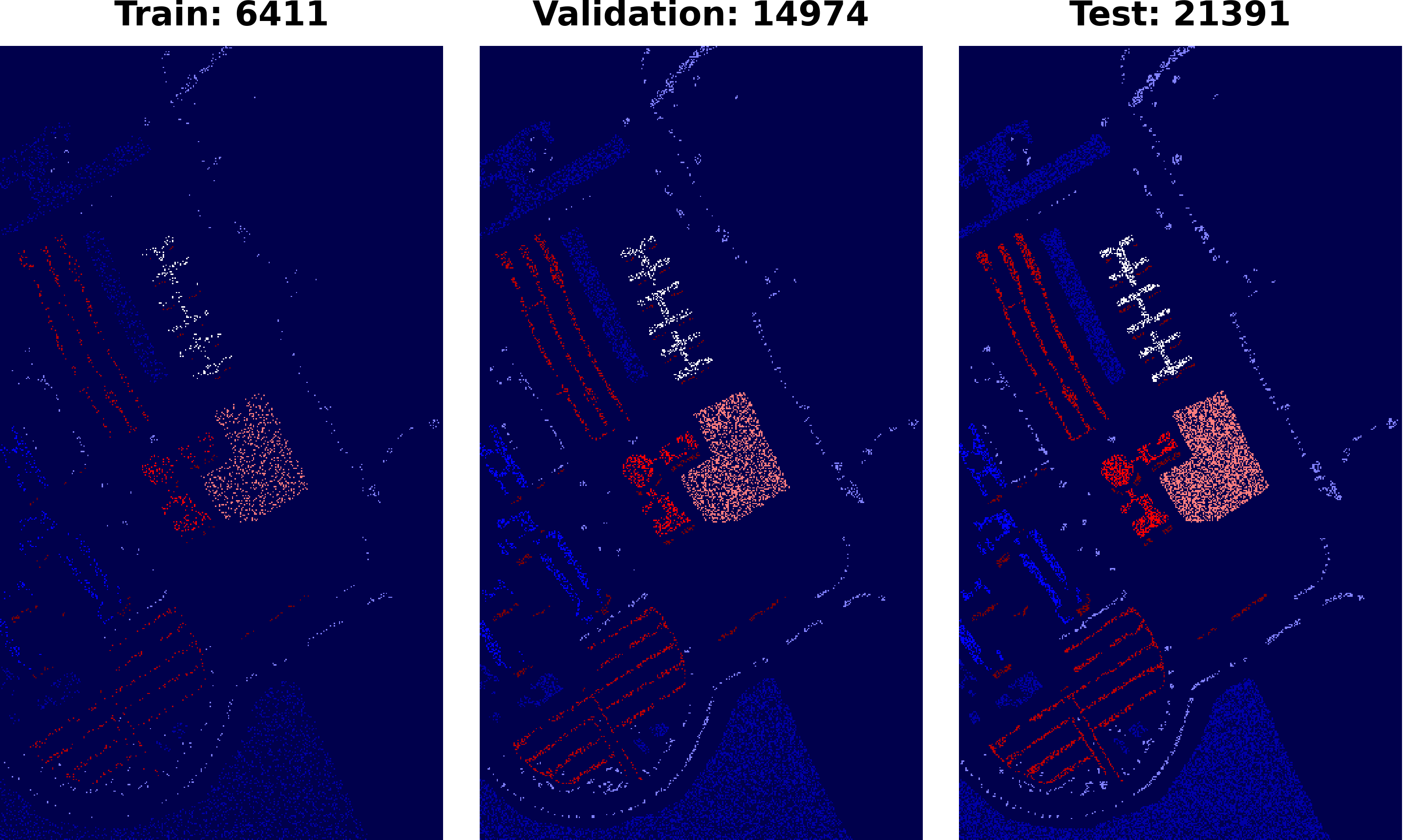}
		\caption{Pavia University (PU) Dataset} 
		\label{Fig2B}
	\end{subfigure}
    \begin{subfigure}{0.48\textwidth}
		\centering
        \includegraphics[width=0.99\textwidth]{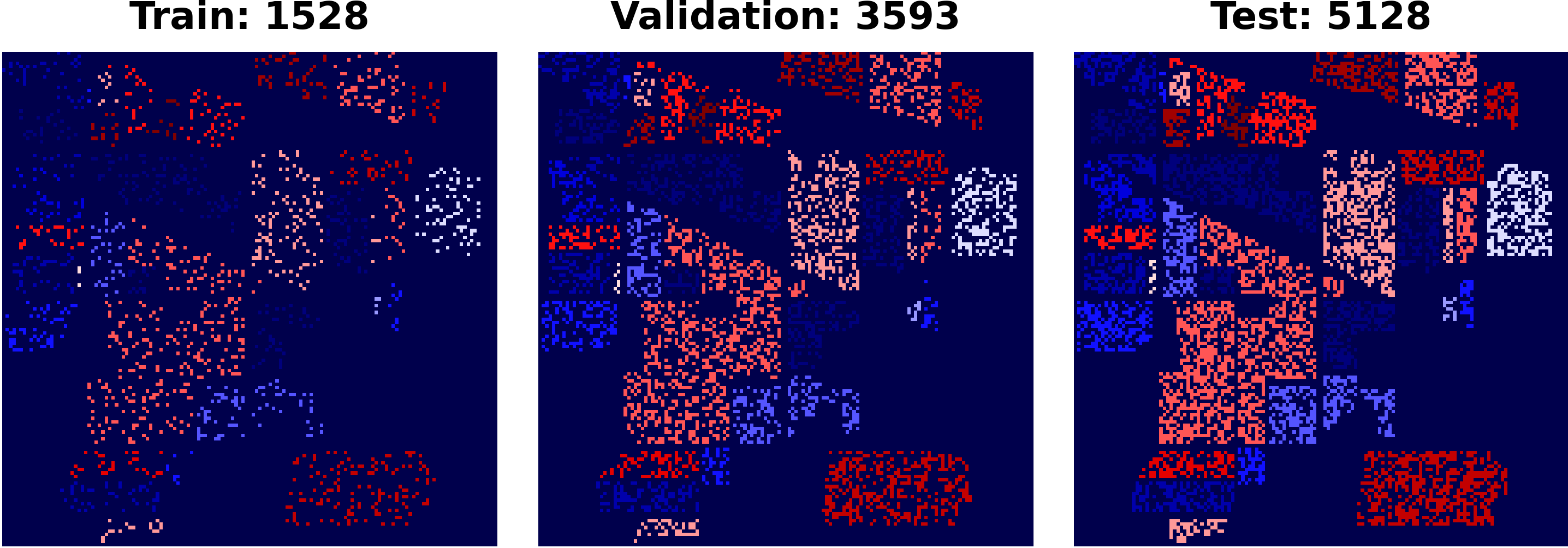}
		\caption{Indian Pines (IP) Dataset} 
		\label{Fig2D}
	\end{subfigure}
\caption{Land cover maps for disjoint Train, Validation, and test Samples. Number of samples are presented in Table \ref{Tab2}.}
\label{Fig2}
\end{figure}

Table \ref{Tab1} outlines the specifics of each dataset employed in the experiments. In contrast, Table \ref{Tab2} furnishes the count of disjoint training, validation, and test samples chosen from each class for the training, validation, and testing phases of both the proposed and comparative methods. Additionally, the geographical maps corresponding to the disjoint training, validation, and test samples are depicted in Figure \ref{Fig2}. It is crucial to emphasize that the consistency in the number of training, validation, and test samples, along with their geographical locations, is maintained across all methods considered in the experimental evaluation. This ensures unbiased and equitable assessments for presenting the results.

\begin{table}[!hbt]
    \centering
    \caption{Overview of HSI Datasets Employed in Experimental Evaluation.}
    \resizebox{\columnwidth}{!}{\begin{tabular}{l|cccc} \hline 
        --- & \textbf{IP} & \textbf{PU} & \textbf{UH} & \textbf{SA}\\  \hline 
        \textbf{Source} & AVIRIS & ROSIS-03 & CASI & AVIRIS \\
        \textbf{Sensor} & Aerial & Aerial & Aerial & Aerial \\
        \textbf{Resolution} & $20~m$ & $1.3~m$ & $2.5~mpp$ & $3.7~m$ \\ 
        \textbf{Wavelength} & $400-2500$ & $430-860$ & $0.35-1.05$ & $0.35-1.05$ \\
        \textbf{Spectral} & 220 & 115 & 144 & 224 \\
        \textbf{Spatial} & $145\times 145$ & $610 \times 610$ & $340\times 1905$ & $340\times 1905$ \\
        \textbf{Samples} & 21025 & 207400 & 1329690 & 54129 \\
        \textbf{Classes} & 16 & 9 & 15 & 16 \\
        \textbf{Year} & 1992 & 2001 & 2013 & ---- \\ \hline 
    \end{tabular}}
    \label{Tab1}
\end{table}
\begin{table*}[!hbt]
    \centering
    \caption{Number of Disjoint Training, Validation, and Test Samples Employed for Training and Evaluating the Proposed and Comparative Models.}
    \begin{tabular}{cc|cc|cc|cc} \hline 
    \multicolumn{2}{c|}{\textbf{Indian Pines}} &  \multicolumn{2}{c|}{\textbf{Salinas}} & \multicolumn{2}{c|}{\textbf{University of Houston}} & \multicolumn{2}{c}{\textbf{Pavia University}} \\ \hline 
    \textbf{Class} & \textbf{Train/Val/Test} & \textbf{Class} & \textbf{Train/Val/Test} & \textbf{Class} & \textbf{Train/Val/Test} & \textbf{Class} & \textbf{Train/Val/Test} \\ \hline 
        Alfalfa & 6/17/23 & Weeds 1 & 301/703/1005 & Healthy Grass & 187/438/626 & Asphalt & 994/2321/3316 \\
        Corn Notill & 214/500/714 & Weeds 2 & 558/1305/1863 & Stressed Grass & 188/439/627 & Meadows & 2797/6527/9325 \\
        Corn Mintill & 124/291/415 & Fallow & 296/692/988 & Synthetic Grass & 104/244/349 & Gravel & 314/735/1050 \\
        Corn & 35/83/119 & Fallow Rough Plow & 209/488/697 & Trees & 186/436/622 & Trees & 459/1073/1532 \\
        Grass Pasture & 72/169/242 & Fallow Smooth & 401/938/1339 & Soil & 186/435/621 & Painted & 201/471/673 \\
        Grass Trees & 109/256/365 & Stubble & 593/1386/1980 & Water & 48/114/163 & Soil & 754/1760/2515 \\
        Grass Mowed & 4/10/14 & Celery & 536/1253/1790 & Residential & 190/444/634 & Bitumen & 199/466/665 \\
        Hay Windrowed & 71/168/239 & Grapes Untrained & 1690/3945/5636 & Commercial & 186/436/622 & Bricks & 552/1289/1841 \\
        Oats & 3/7/10 & Soil Vinyard Develop & 930/2171/3102 & Road & 187/439/626 & Shadows & 141/332/474 \\
        Soybean Notill & 145/341/486 & Corn Weeds & 491/1148/1639 & Highway & 183/430/614 &  &  \\
        Soybean Mintill & 368/859/1228 & Lettuce 4wk & 160/374/534 & Railway & 185/432/618 &  &  \\
        Soybean Clean & 88/208/297 & Lettuce 5wk & 288/675/964 & Parking Lot 1 & 184/432/617 &  &  \\
        Wheat & 30/72/103 & Lettuce 6wk & 137/321/458 & Parking Lot 2 & 70/164/235 &  &  \\
        Woods & 189/443/633 & Lettuce 7wk & 160/375/535 & Tennis Court & 64/150/214 &  &  \\
        Buildings & 57/136/193 & Vinyard Untrained & 1090/2544/3634 & Running Track & 99/231/330 &  &  \\
        Stone Steel & 13/33/47 & Vinyard Trellis & 270/633/904 &  &  &  &  \\ \hline 
    \end{tabular}
    \label{Tab2}
\end{table*}

The \textbf{Salinas (SA)} data cube was captured by the Airborne Visible/Infrared Imaging Spectrometer (AVIRIS) over Salinas Valley, California. This cube comprises 224 bands and boasts a high spatial resolution of 3.7-meter pixels. The spatial dimensions encompass a total of $512 \times 217$ samples. In preparation for the experiments, the 20 most noise-prone and water-absorbing bands, specifically [108-112], [154-167], and 224, were excluded. The SA cube, available solely as at-sensor radiance data, encompasses diverse features such as bare soils, vineyard fields, and vegetables. Notably, the SA cube encompasses samples across 16 distinct classes, forming the ground truths for the dataset.

The \textbf{Pavia University (PU)} data cube, captured by the Reflective Optics System Imaging Spectrometer (ROSIS) sensor during a flight campaign over Pavia in Northern Italy, boasts a geometric resolution of 1.3 meters. Comprising 103 spectral bands and featuring spatial dimensions of $610 \times 610$ lines (spatial pixels), it is worth noting that certain samples lack information and must be excluded prior to experiments. The PU cube encompasses samples across 9 distinct classes, forming the ground truths for the dataset.

The \textbf{Indian Pines (IP)} dataset was collected by the Airborne Visible/Infrared Imaging Spectrometer (AVIRIS) over the Indian Pines test site in North-western Indiana \cite{green1998imaging}. It encompasses $224$ spectral bands spanning a wavelength range from $400$ to $2500$ $nm$. To enhance data quality, $24$ null and corrupted bands were excluded. The spatial dimensions of the image are $145\times{145}$ pixels, representing $16$ distinct and mutually exclusive vegetation classes. The spatial resolution is 20 meters per pixel (MPP).

The \textbf{University of Houston} (UH) dataset, published by the IEEE Geoscience and Remote Sensing Society as part of the 2013 Data Fusion Contest, was gathered by the Compact Airborne Spectrographic Imager (CASI). With a spatial dimension of $340\times{1905}$ pixels and 144 spectral bands, this dataset exhibits a spatial resolution of $2.5$ meters per pixel (MPP) and a wavelength range from $0.38$ to $1.05$ $\mu$m. Notably, the ground truth for this dataset encompasses 15 distinct land-cover classes.

\subsection{Experimental Settings}

Research-focused publications extensively conduct experimental evaluations to elucidate the merits and drawbacks of proposed methodologies. However, inconsistencies in experimental settings, particularly in the random selection of training, validation, and test samples, may impede fair comparisons among different works. To ensure fairness, it is imperative to maintain identical experimental settings, including consistent geographical locations for chosen models and uniform sample numbers for each training round in cross-validation. Typically, random sample selection introduces variability, potentially leading to differences among models executed at different times.

Another prevalent issue in recent literature is the overlapping of training/test samples. In some cases, random selection for training and validation includes the entire dataset during testing, resulting in a biased model with inflated accuracy. To address this, in our work, while training/test samples are randomly chosen concurrently for all models, special attention is paid to ensuring an empty intersection among these samples, mitigating biases introduced by overlapping.

Various metrics can assess the performance of a classification model, including Overall Accuracy (OA), Average Accuracy (AA), and the Kappa ($\kappa$) coefficient, alongside other statistical tests. OA, expressed as a percentage, provides insights into correctly mapped samples, offering basic classification information that is easy to compute and understand. On the other hand, the $\kappa$ coefficient, derived from statistical tests, gauges classification accuracy by comparing model performance to random values. Ranging from -1 to 1, where -1, 0, and 1 indicate significantly worse, equal to, or better than random classification, the $\kappa$ coefficient is calculated as follows:

\begin{equation}
    \kappa = \frac{p_o-p_e}{1-p_e}
\end{equation}

In this context, $p_o$ represents the OA, and $p_e$ characterizes the measures of agreement between actual and predicted class labels occurring by chance. Additionally, the difference $p_o - p_e$ captures the variance between the observed OA accuracy of the model and the OA accuracy expected by chance. The term $1 - p_e$ calculates the maximum value for this difference. For a model to be deemed effective, the maximum and observed differences should be close, resulting in $\kappa = 1$. Conversely, for a random model, the numerator becomes 0, leading to $\kappa = 0$ or potentially a negative value. In such cases, the OA accuracy of the model falls below what could have been achieved by a random guess.

In our experimental setup, the assessment of both the proposed and comparative models began with the utilization of 5\% randomly selected samples for training and 45\% for validation. A mini-batch size of 56, the Adam optimizer, and specific learning parameters, including a learning rate of 0.0001 and a decay rate of 1e-06, were employed over 50 epochs. These experiments were initialized using an $8 \times 8$ patch size as a foundational configuration. Furthermore, to comprehensively assess the performance of the proposed model and compare it with other methods, various configurations involving different numbers of training samples and patch sizes were systematically tested in our experiments. This exploration aimed to provide a thorough understanding of how the model performs under diverse training scenarios and spatial resolutions.

\subsection{Effects of Patch Sizes, Training Samples, and Attentional Heads on Model Performance}

This section focuses on three key factors that can significantly impact the performance of this fusion model: patch sizes, training samples, and attentional heads. The effects of patch sizes and training samples on the performance of a model, particularly one that incorporates the attentional fusion of 3D ST and SST, are crucial considerations in HSIC. This discussion explores the impact of varying patch sizes and training samples on the model's ability to capture spatial and spectral information, thereby influencing its overall classification performance.

\begin{figure}[!hbt]
    \centering
	\begin{tikzpicture}[scale=0.90]
    \begin{axis}[xlabel={Patch Size}, ylabel={Performance ($\kappa$)}, xmin=2, xmax=10, ymin=80, ymax=100, xtick={2,4,6,8,10}, ytick={80,82,84,86,88,90,92,94,96,98,100}, legend pos=south east, xmajorgrids=true, ymajorgrids=true, grid style=dashed]
        \addplot[draw=blue, mark=*]
        coordinates{(2,96.23)(4,97.27)(6,97.84)(8,97.94)(10,98.17)};
        \addplot[color=black, mark=triangle*]
        coordinates{(2,94.44)(4,96.89)(6,97.6)(8,97.64)(10,97.86)};
        \addplot[color=red, mark=square]
        coordinates{(2,82.13)(4,84.85)(6,87.64)(8,88.79)(10,89.72)};
        \addplot[style=dashed, color=black, mark=o]
        coordinates{(2,95.2)(4,97.48)(6,98.04)(8,98.6)(10,98.92)};
        \legend{University of Houston, Pavia University, Indian Pines, Salinas}
    \end{axis}
    \end{tikzpicture}
	\caption{Impact of Patch Sizes ($2 \times 2$, $4 \times 4$, $6 \times 6$, $8 \times 8$, and $10 \times 10$) on Model Performance. In all experiments, 50\% of the samples are designated for testing, while the remaining 50\% are further divided into 95\% for validation and 5\% for training.}
    \label{Fig3}
\end{figure}
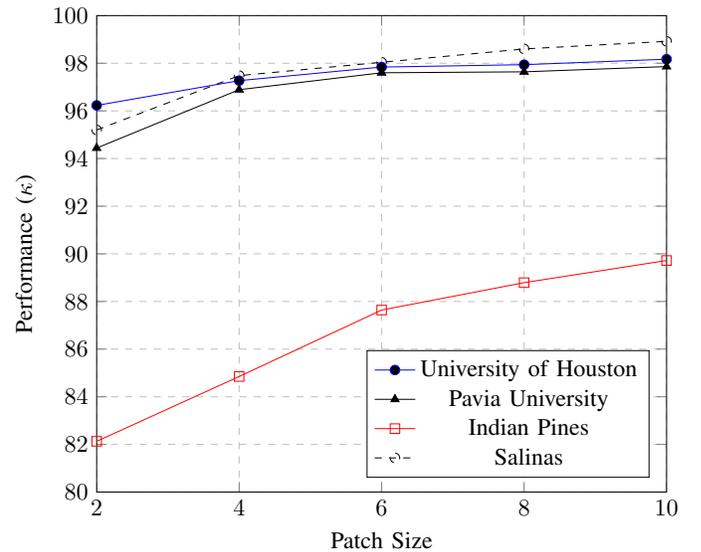

Patch size refers to the spatial extent of the input patches fed into the fusion model. It plays a crucial role in capturing local spatial information and contextual relationships within the HSI data. The choice of patch size can affect the model's ability to capture fine-grained spatial details and contextual dependencies. Increasing patch sizes may enhance the model's ability to capture global spatial relationships within the HSIs. Larger patches provide a broader contextual understanding of the spatial layout of different features, contributing to improved spatial feature extraction. Moreover, larger patches allow the model to incorporate more spectral information within each token, potentially aiding in the extraction of complex spectral signatures. However, this might lead to challenges in capturing fine-grained spectral details. Conversely, smaller patch sizes may focus on local details, capturing finer spatial structures. This could be advantageous when dealing with intricate patterns or objects with distinct spatial characteristics. Furthermore, smaller patches might better capture specific spectral characteristics and variations, enabling the model to discern subtle differences between classes. However, this may limit the model's ability to handle complex spectral patterns. Therefore, an optimal patch size strikes a balance between capturing spatial details and contextual information, leading to improved classification accuracy. The impact of various patch sizes on the proposed model is illustrated in Figure \ref{Fig3}.

Moreover, in scenarios where labeled training samples are scarce, the model might struggle to generalize well to unseen data. insufficient training samples may result in underfitting, where the model fails to capture the underlying patterns in the data. Furthermore, affecting the model's robustness and generalization to diverse HSI scenes which limits the model's ability to generalize and leads to reduced classification accuracy. Moreover, It is crucial to have an adequately sized training set that covers diverse spectral signatures and representative samples from each class. Increasing the number of labeled training samples provides the model with more diverse examples, facilitating better learning and generalization. 

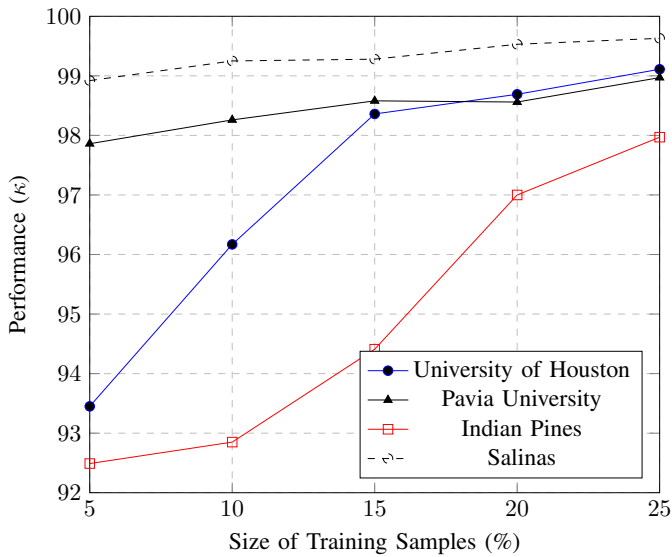
\begin{figure}[!hbt]
    \centering
		\begin{tikzpicture}[scale=0.90]
        \begin{axis}[
        xlabel={Size of Training Samples (\%)},
        ylabel={Performance ($\kappa$)},
        xmin=5, xmax=25, ymin=92, ymax=100,
        xtick={5,10,15,20,25},
        ytick={92,93,94,95,96,97,98,99,100},
        legend pos=south east, xmajorgrids=true, ymajorgrids=true, grid style=dashed,]
        \addplot[draw=blue, mark=*]
        coordinates{(5,93.45)(10,96.17)(15,98.36)(20,98.69)(25,99.11)};
        \addplot[color=black, mark=triangle*]
        coordinates{(5,97.86)(10,98.26)(15,98.58)(20,98.56)(25,98.97)};
        \addplot[color=red, mark=square]
        coordinates{(5,92.49)(10,92.85)(15,94.41)(20,97.0)(25,97.97)};
        \addplot[style=dashed, color=black, mark=o]
        coordinates{(5,98.92)(10,99.25)(15,99.28)(20,99.53)(25,99.63)};
        \legend{University of Houston, Pavia University, Indian Pines, Salinas}
        \end{axis}
        \end{tikzpicture}
		\caption{Impact of Training Samples on Model Performance. In all experiments, a $10 \times 10$ patch size is employed, yielding optimal results, as depicted in Figure \ref{Fig3}.}
    \label{Fig4}
\end{figure}

Adequate samples contribute to improved model performance, especially in handling variations within different classes. In another scenario, the distribution of training samples across different classes can also impact model performance. Class imbalance, where one or more classes have significantly fewer samples than others, can lead to biased models that favor dominant classes. It is important to ensure a balanced distribution of training samples to prevent such biases and improve the model's ability to generalize to all classes. The impact of various sample sizes on the proposed model is illustrated in Figure \ref{Fig4}.

\begin{figure}[!hbt]
    \centering
		\begin{tikzpicture}[scale=0.90]
        \begin{axis}[
        xlabel={Number of Attentional Heads},
        ylabel={Performance ($\kappa$)},
        xmin=2, xmax=12, ymin=96, ymax=100,
        xtick={2,4,6,8,10,12},
        ytick={96,97,98,99,100},
        legend pos=south east, xmajorgrids=true, ymajorgrids=true, grid style=dashed,]
        \addplot[draw=blue, mark=*]
        coordinates{(2,99.24)(4,99.05)(6,99.15)(8,99.32)(10,99.21)(12,99.11)};
        \addplot[color=black, mark=triangle*]
        coordinates{(2,98.53)(4,98.44)(6,98.61)(8,98.97)(10,98.48)(12,98.66)};
        \addplot[color=red, mark=square]
        coordinates{(2,96.72)(4,97.9)(6,98.12)(8,98.61)(10,97.84)(12,97.94)};
        \addplot[style=dashed, color=black, mark=o]
        coordinates{(2,99.4)(4,99.48)(6,99.30)(8,99.33)(10,99.42)(12,99.43)};
        \legend{University of Houston, Pavia University, Indian Pines, Salinas}
        \end{axis}
        \end{tikzpicture}
		\caption{Impact of Number of Attentional Heads on Model Performance. In all experiments, a $10 \times 10$ patch size and 25\% training and validation samples, and the remaining 50\% samples used for the test are employed, yielding optimal results, as depicted in Figures \ref{Fig3} and \ref{Fig4}.}
    \label{Fig5}
\end{figure}
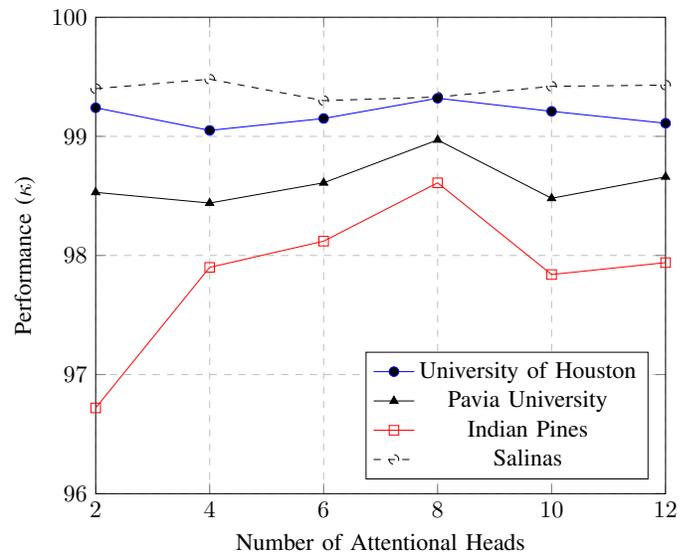

To determine the most effective number of attention heads, we investigated the influence of varying their quantity on classification accuracy in our experiments. This investigation was carried out while keeping the patch size $10 \times 10$, the number of training, validation, and test samples (25\%, 25\%, and 50\% respectively), and the number of tokens fixed at 64, respectively. The outcomes of these experiments are illustrated in Figure \ref{Fig5}. Considering all four datasets, we ultimately opted for 8 as the optimal number of attention heads.

\subsection{Quantitative and Qualitative Results and Discussion}

HSIC is an important task with applications in various domains. Traditional approaches for HSIC often rely on handcrafted features and 2D CNNs. However, these methods may struggle to capture the spatial-spectral characteristics of HSI data effectively. To address these challenges, recent research has explored the use of Deep Learning models, including 3D CNNs, 2D and 3D Inception Nets, Hybrid CNNs, (2+1)D Extreme Expansion Nets, Attention Graph CNNs, and Transformers. This section will focus on the experimental results of attentional feature fusion of 3D Swin Transformer (3D ST) and Spatial-Spectral Transformer (SST) for HSIC, comparing their performance with the aforementioned models. To evaluate the performance of attentional feature fusion of Transformers for HSIC, a comprehensive experimental setup was employed. The dataset used for evaluation consisted of a large collection of HSIs, with ground truth labels for different classes. The dataset was split into training, validation, and testing sets.

\begin{table*}[!hbt]
    \centering
    \caption{\textbf{Indian Pines:} Per-Class Classification results of proposed method along with the comparative methods. Class-wise ground truth maps are presented in Figure \ref{Fig6}.}
    \resizebox{\textwidth}{!}{\begin{tabular}{c|cc|cc|cc|cc|cc|cc|cc|cc|cc} \hline 
    \multirow{2}{*}{\textbf{Class}} & \multicolumn{2}{c|}{\textbf{2D CNN}} & \multicolumn{2}{c|}{\textbf{3D CNN}} & \multicolumn{2}{c|}{\textbf{Hybrid CNN}} & \multicolumn{2}{c|}{\textbf{Hybrid IN}} & \multicolumn{2}{c|}{\textbf{2D IN}} & \multicolumn{2}{c|}{\textbf{3D IN}} & \multicolumn{2}{c|}{\textbf{(2+1)D XN}} & \multicolumn{2}{c|}{\textbf{Attention GCN}} & \multicolumn{2}{c}{\textbf{Proposed}} \\ \cline{2-19} 
    
    & Val & Test & Val & Test & Val & Test & Val & Test & Val & Test & Val & Test & Val & Test & Val & Test & Val & Test \\ \hline
    
    Alfalfa & 84.61 & 100 & 100 & 100 & 96.15 & 100 & 96.15 & 100 & 92.30 & 100 & 96.15 & 100 & 100 & 100 & 69.23 & 80.00 & 100 & 100 \\
    Corn notill & 91.62 & 89.86 & 98.37 & 97.55 & 89.5 & 91.60 & 99.625 & 98.95 & 94.87 & 97.90 & 98.50 & 97.90 & 97.25 & 96.50 & 81.37 & 81.81 & 99.50 & 99.30 \\
    Corn mintill & 86.23 & 79.51 & 96.98 & 98.79 & 87.09 & 92.16 & 99.56 & 99.39 & 96.55 & 99.39 & 98.70 & 98.79 & 96.34 & 97.59 & 83.87 & 89.15 & 98.06 & 98.79 \\ 
    Corn & 84.96 & 83.33 & 93.98 & 100 & 69.92 & 81.25 & 99.24 & 100 & 95.48 & 93.75 & 97.74 & 97.91 & 91.72 & 95.83 & 64.66 & 60.41 & 99.24 & 100 \\ 
    Grass pasture & 91.88 & 90.72 & 98.89 & 98.96 & 97.04 & 98.96 & 100 & 98.96 & 98.52 & 97.93 & 100 & 100 & 96.30 & 98.96 & 91.14 & 94.84 & 99.26 & 98.96 \\ 
    Grass trees & 99.02 & 99.31 & 99.75 & 99.31 & 99.75 & 99.31 & 99.75 & 100 & 99.51 & 98.63 & 100 & 100 & 99.75 & 98.63 & 97.79 & 98.63 & 100 & 100 \\ 
    Grass mowed & 87.5 & 100 & 81.25 & 100 & 68.75 & 83.33 & --- & --- & 87.5 & 100 & 68.75 & 83.33 & 68.75 & 83.33 & 62.50 & 83.33 & 100 & 100 \\ 
    Hay windrowed & 100 & 100 & 100 & 100 & 100 & 100 & 100 & 100 & 100 & 100 & 100 & 100 & 100 & 100 & 98.88 & 98.95 & 100 & 100 \\ 
    Oats & 16.66 & --- & 83.33 & 100 & 50 & 50 & --- & --- & 83.33 & 100 & 50 & 50 & 91.66 & 75.00 & 16.66 & --- & 100 & 100 \\ 
    Soybean notill & 85.29 & 87.69 & 94.11 & 91.79 & 86.58 & 89.23 & 95.40 & 97.43 & 91.36 & 93.84 & 94.85 & 95.89 & 92.64 & 92.82 & 90.62 & 90.76 & 95.58 & 95.38 \\ 
    Soybean mintill & 96.94 & 97.35 & 98.69 & 99.18 & 93.74 & 95.11 & 98.61 & 98.37 & 96.43 & 97.35 & 98.61 & 99.18 & 97.74 & 96.94 & 92.14 & 95.11 & 99.34 & 99.38 \\ 
    Soybean clean & 89.75 & 92.43 & 97.28 & 97.47 & 88.55 & 91.59 & 99.09 & 99.15 & 96.38 & 97.47 & 99.09 & 99.15 & 96.98 & 94.95 & 89.15 & 88.23 & 99.39 & 100 \\ 
    Wheat & 99.13 & 97.56 & 100 & 100 & 100 & 100 & 100 & 100 & 100 & 100 & 100 & 100 & 100 & 100 & 98.26 & 97.56 & 100 & 100 \\ 
    Woods & 98.87 & 100 & 99.43 & 100 & 97.60 & 98.41 & 99.57 & 100 & 98.73 & 98.41 & 99.71 & 100 & 98.30 & 99.60 & 96.33 & 98.02 & 99.85 & 99.60 \\ 
    Buildings & 93.05 & 94.87 & 98.14 & 100 & 98.14 & 100 & 99.53 & 100 & 96.29 & 97.43 & 100 & 100 & 96.29 & 100 & 86.11 & 93.58 & 98.61 & 100 \\ 
    Stone Steel & 98.07 & 100 & 100 & 100 & 98.07 & 100 & 100 & 100 & 100 & 100 & 98.07 & 100 & 100 & 100 & 88.46 & 89.47 & 100 & 100 \\ \hline 

    \textbf{Train (S)} & \multicolumn{2}{c|}{41.62} & \multicolumn{2}{c|}{69.58} & \multicolumn{2}{c|}{83.67} & \multicolumn{2}{c|}{204.44} & \multicolumn{2}{c|}{34.72} & \multicolumn{2}{c|}{323.30} & \multicolumn{2}{c|}{587.05} & \multicolumn{2}{c|}{86.89} & \multicolumn{2}{c}{150.78} \\
    \textbf{Time (S)} & 1.41 & 0.36 & 0.74 & 0.27 & 1.49 & 0.32 & 1.85 & 0.53 & 1.13 & 0.68 & 5.30 & 1.33 & 8.93 & 3.28 & 1.77 & 0.68 & 3.27 & 1.35 \\
    \textbf{Kappa} & 92.55 & 92.43 & 97.80 & 98.11 & 91.59 & 93.90 & 98.27 & 98.49 & 96.03 & 97.34 & 98.29 & 98.61 & 96.67 & 96.84 & 88.28 & 90.39 & 98.91 & 99.00 \\
    \textbf{OA} & 93.49 & 93.38 & 98.07 & 98.35 & 92.62 & 94.65 & 86.66 & 98.39 & 96.52 & 97.66 & 98.50 & 98.78 & 97.07 & 97.23 & 89.73 & 91.58 & 99.03 & 99.11 \\
    \textbf{AA} & 87.72 & 88.29 & 96.27 & 98.94 & 88.81 & 91.94 & 98.59 & 87.02 & 95.46 & 98.26 & 93.76 & 95.14 & 95.24 & 95.64 & 81.69 & 83.75 & 99.30 & 99.47 \\ \hline 
    \end{tabular}}
    \label{Tab3}
\end{table*}
\begin{figure*}[!hbt]
    \centering
	\begin{subfigure}{0.32\textwidth}
		\includegraphics[width=0.99\textwidth]{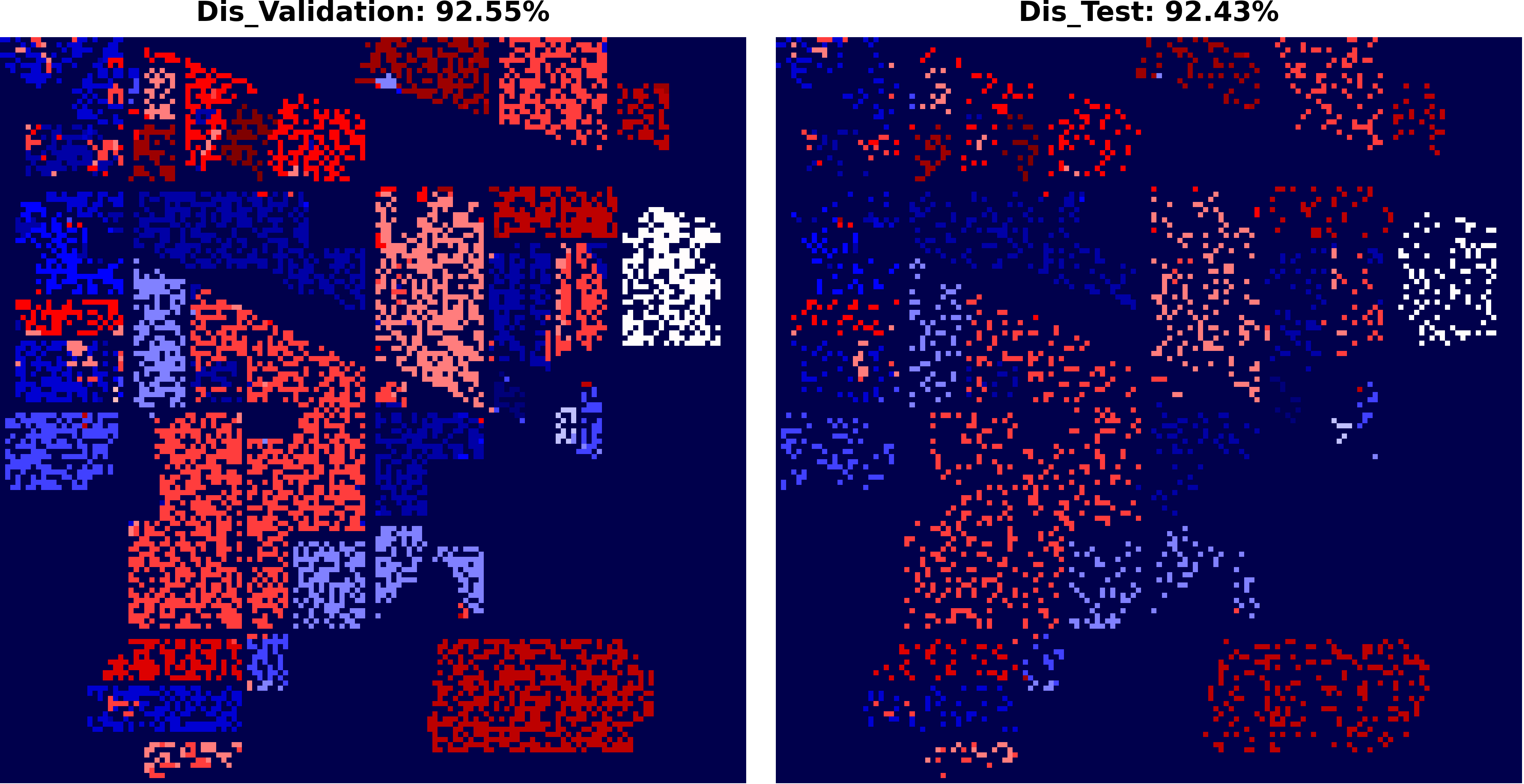}
		\caption{2D CNN} 
		\label{Fig6A}
	\end{subfigure}
	\begin{subfigure}{0.32\textwidth}
		\includegraphics[width=0.99\textwidth]{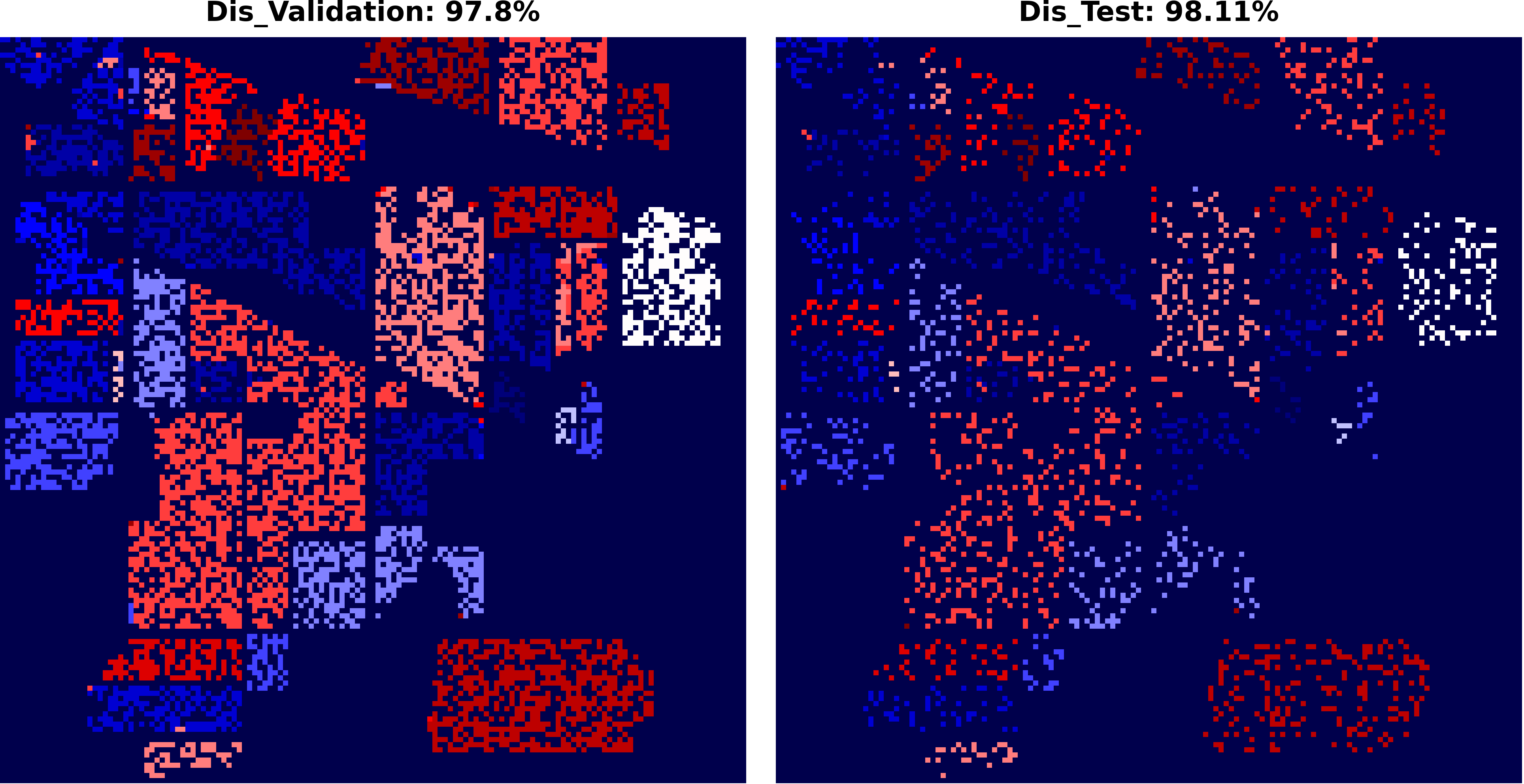}
		\caption{3D CNN}
		\label{Fig6B}
	\end{subfigure} 
	\begin{subfigure}{0.32\textwidth}
		\includegraphics[width=0.99\textwidth]{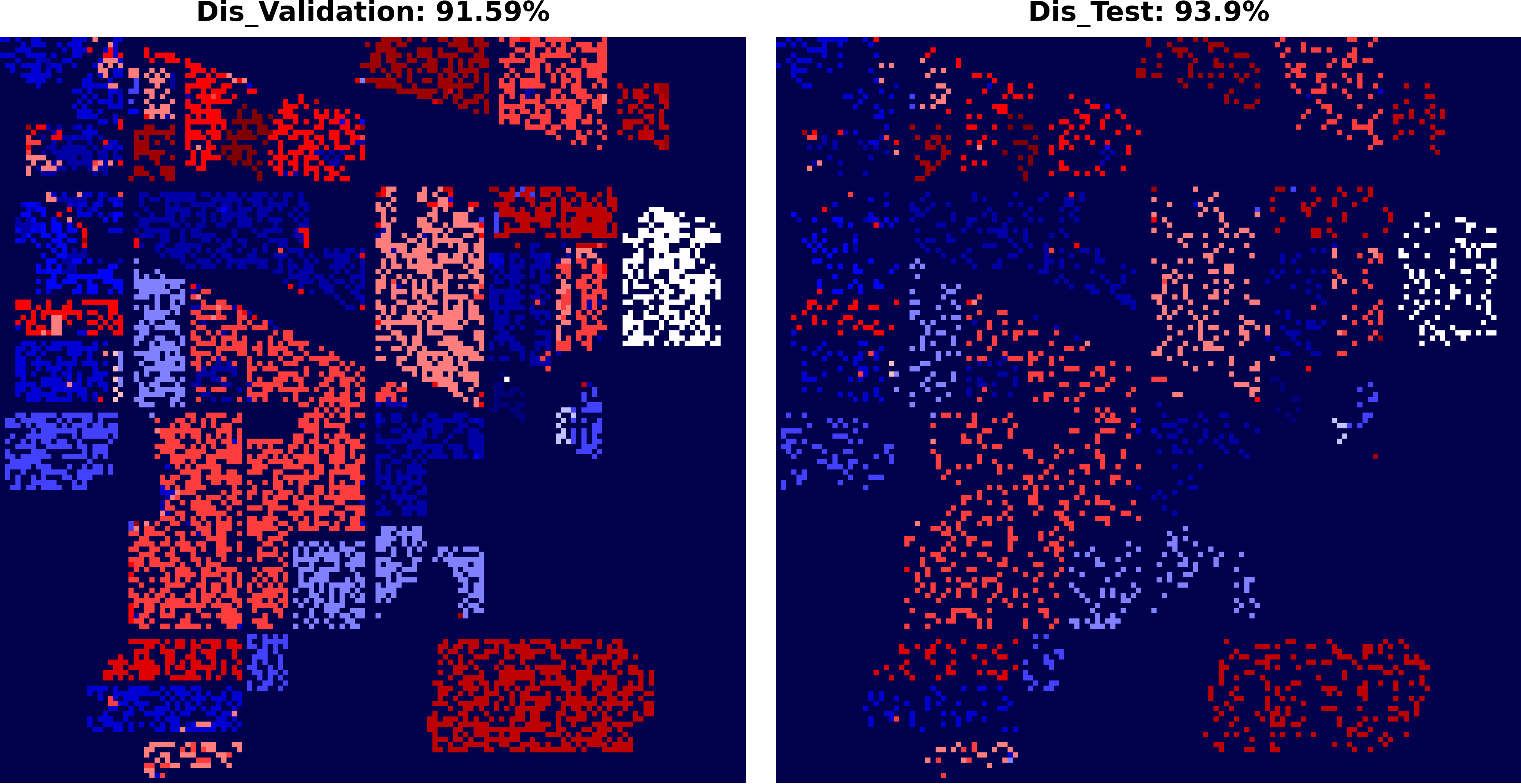}
		\caption{Hybrid CNN}
		\label{Fig6C}
	\end{subfigure} 
	\begin{subfigure}{0.32\textwidth}
		\includegraphics[width=0.99\textwidth]{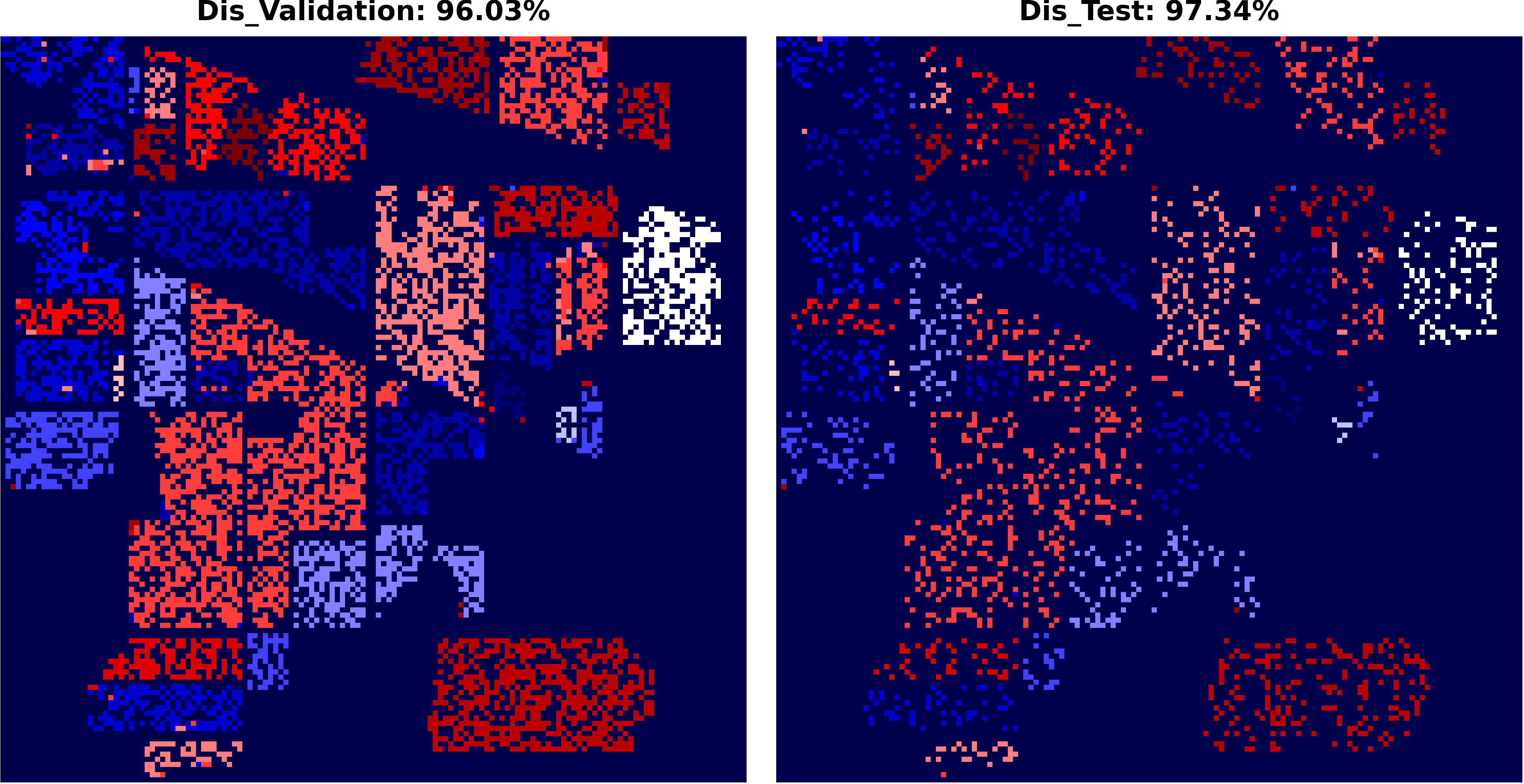}
		\caption{2D Inception Net (IN)}
		\label{Fig6D}
	\end{subfigure} 
	\begin{subfigure}{0.32\textwidth}
		\includegraphics[width=0.99\textwidth]{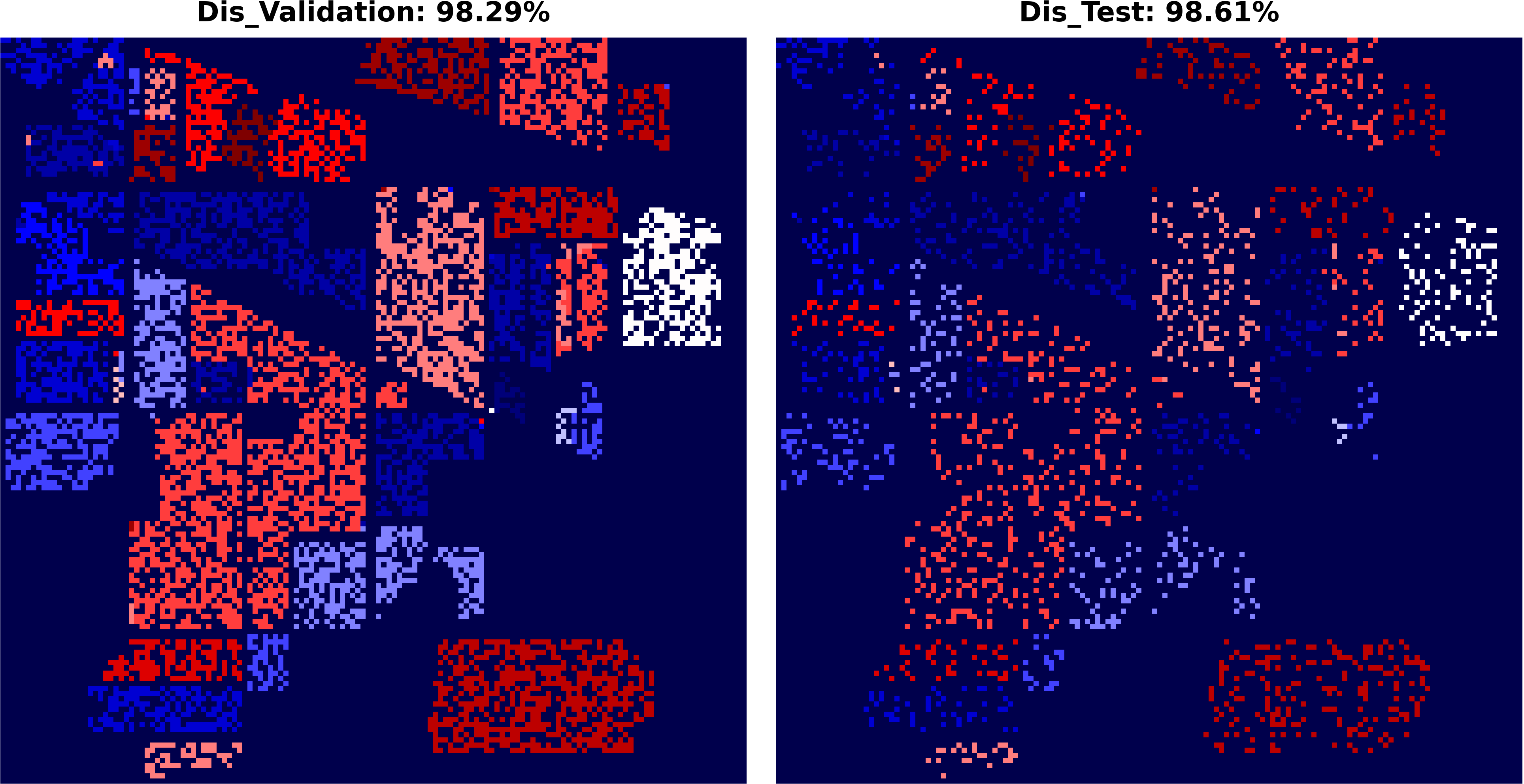}
		\caption{3D Inception Net (IN)}
		\label{Fig6E}
	\end{subfigure} 
	\begin{subfigure}{0.32\textwidth}
		\includegraphics[width=0.99\textwidth]{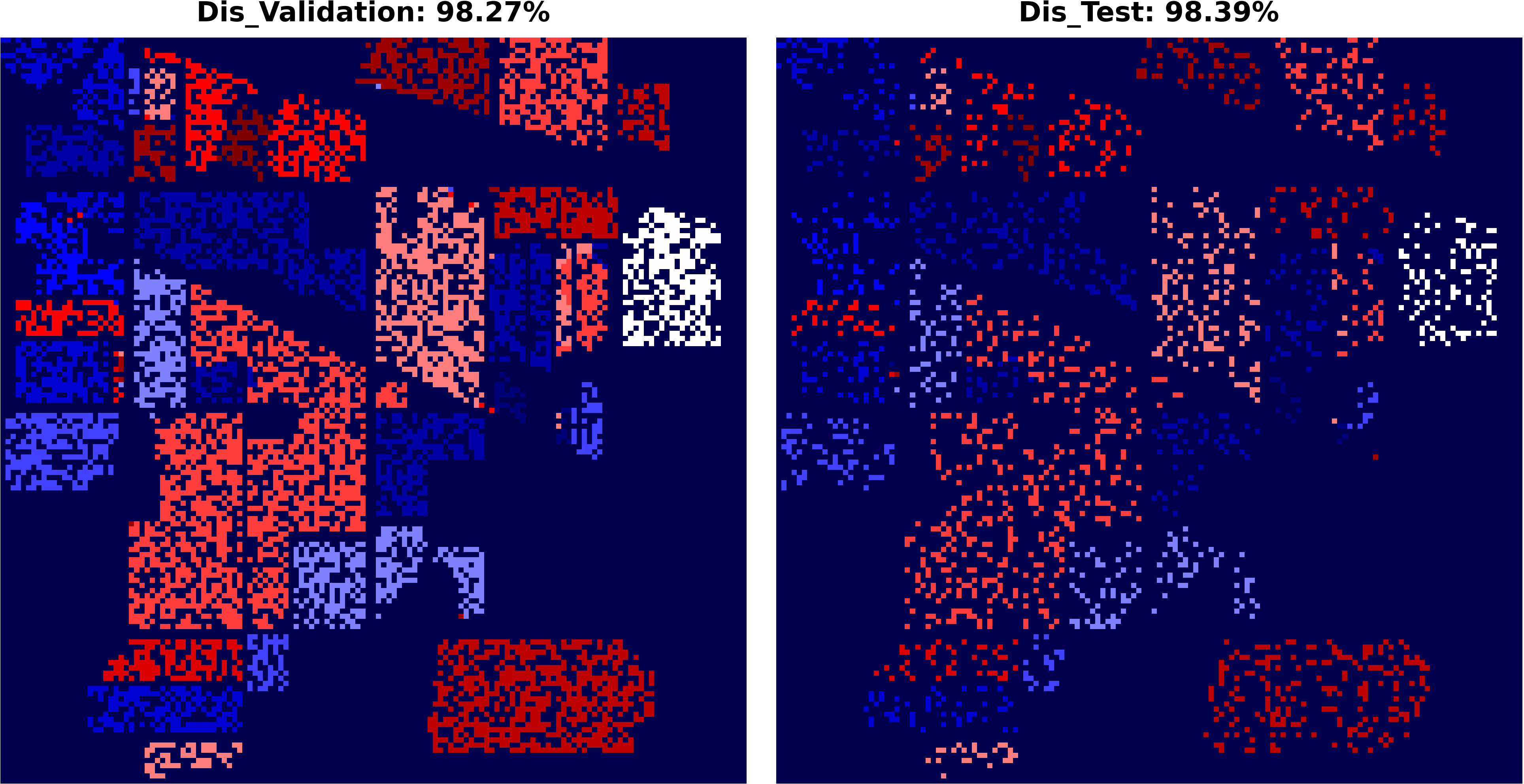}
		\caption{Hybrid Inception Net (IN)}
		\label{Fig6F}
	\end{subfigure}
    \begin{subfigure}{0.32\textwidth}
		\includegraphics[width=0.99\textwidth]{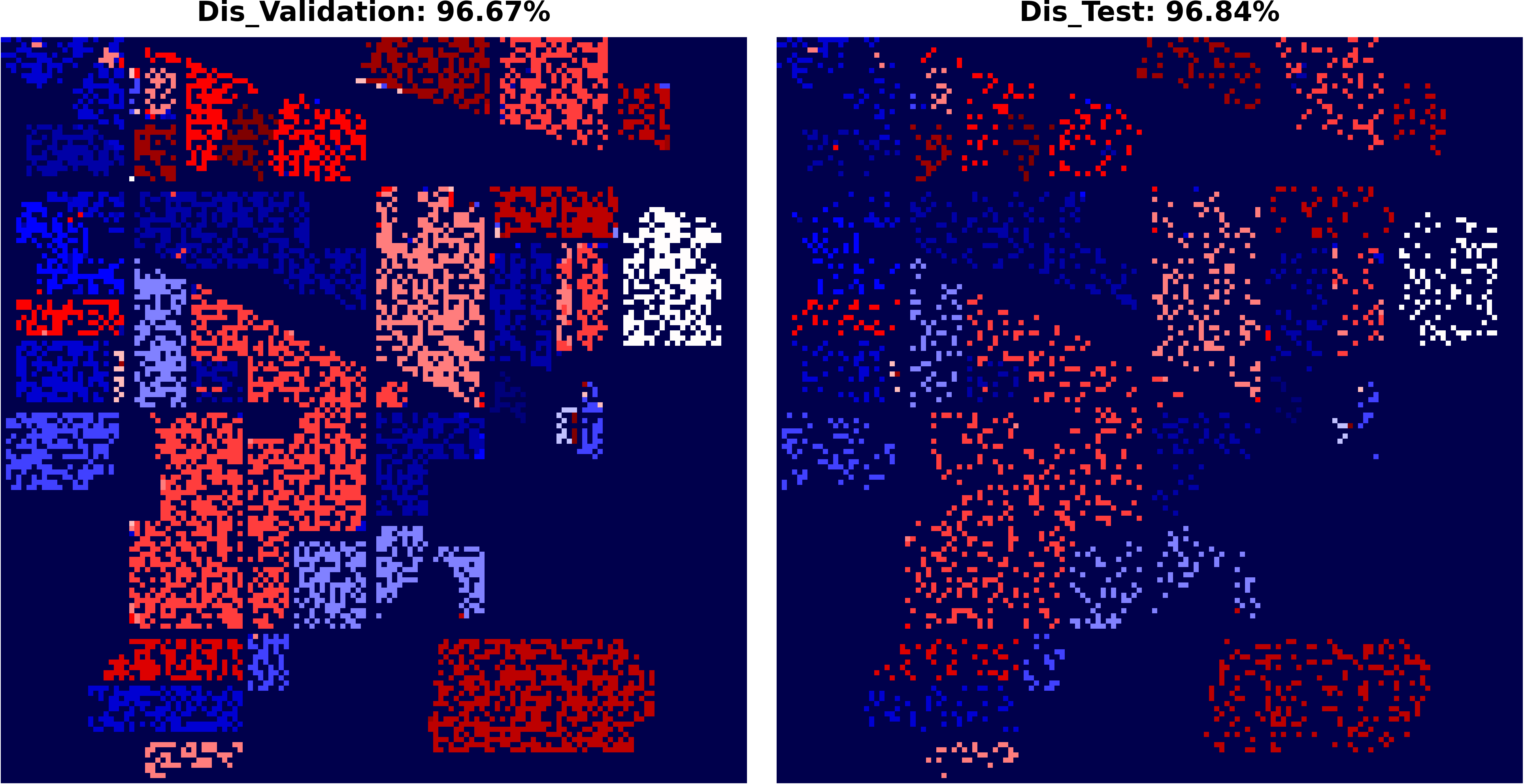}
		\caption{(2+1)D Extreme Xception Net (XN)}
		\label{Fig6G}
	\end{subfigure} 
	\begin{subfigure}{0.32\textwidth}
		\includegraphics[width=0.99\textwidth]{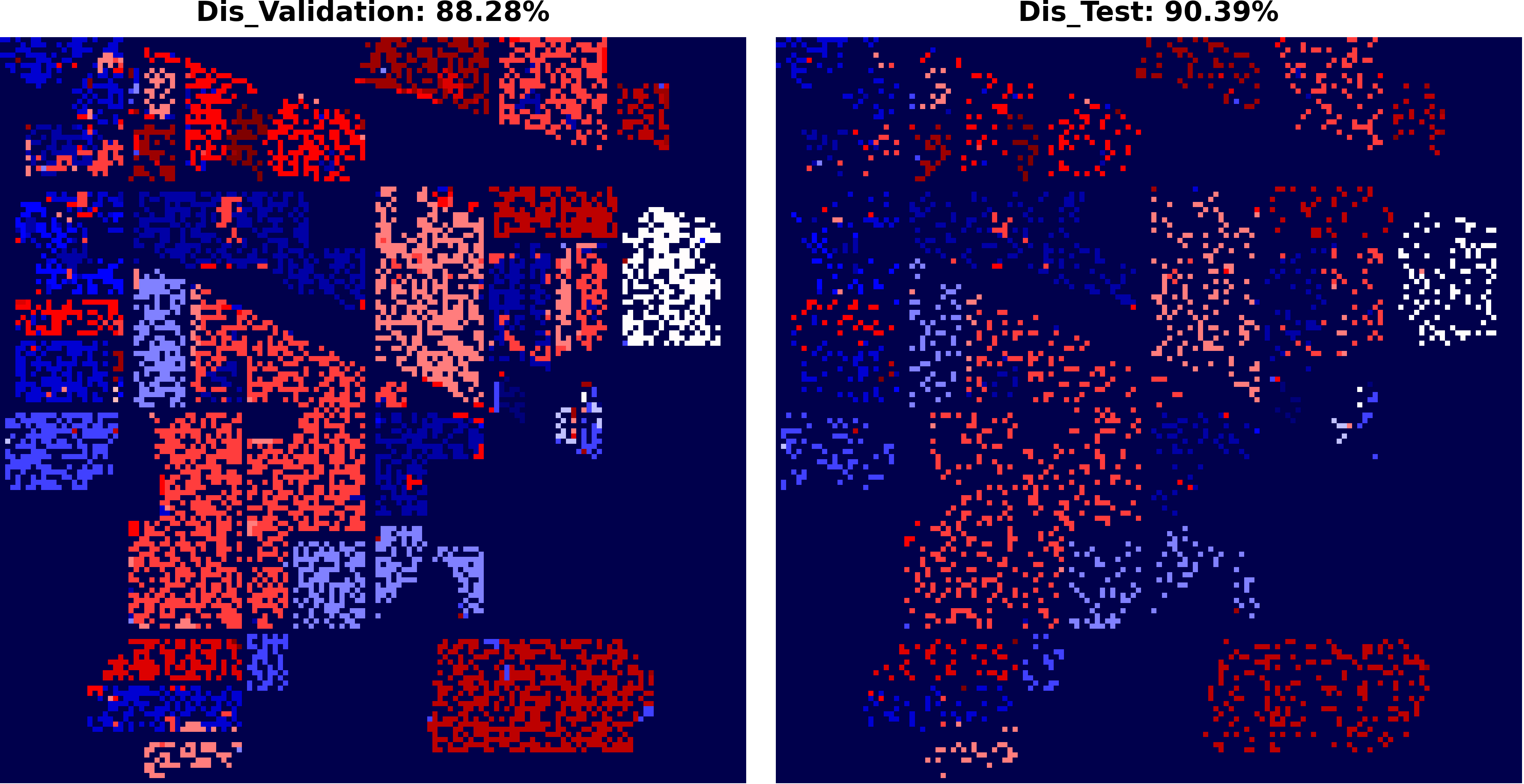}
		\caption{Attention GCN}
		\label{Fig6H}
	\end{subfigure} 
	\begin{subfigure}{0.32\textwidth}
		\includegraphics[width=0.99\textwidth]{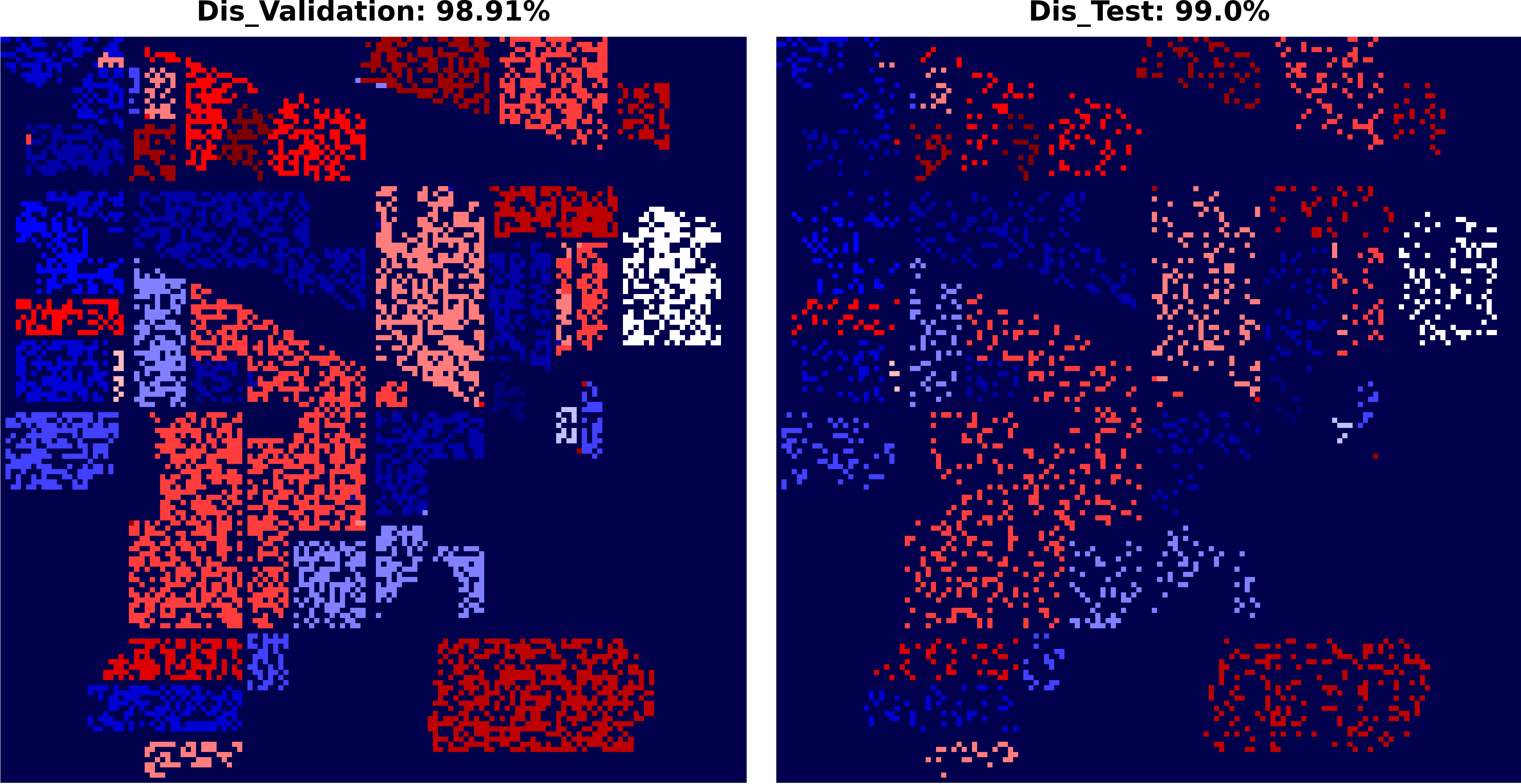}
		\caption{Proposed}
		\label{Fig6I}
	\end{subfigure} 
\caption{\textbf{Indian Pines Dataset:} Land cover maps for disjoint validation and test set are provided. Comprehensive class-wise results can be found in Table \ref{Tab3}.}
\label{Fig6}
\end{figure*}

\begin{table*}[!hbt]
    \centering
    \caption{\textbf{Pavia University:} Per-Class Classification results of proposed method along with the comparative methods. Class-wise ground truth maps are presented in Figure \ref{Fig7}.}
    \resizebox{\textwidth}{!}{\begin{tabular}{c|cc|cc|cc|cc|cc|cc|cc|cc|cc} \hline 
    \multirow{2}{*}{\textbf{Class}} & \multicolumn{2}{c|}{\textbf{2D CNN}} & \multicolumn{2}{c|}{\textbf{3D CNN}} & \multicolumn{2}{c|}{\textbf{Hybrid CNN}} & \multicolumn{2}{c|}{\textbf{Hybrid IN}} & \multicolumn{2}{c|}{\textbf{2D IN}} & \multicolumn{2}{c|}{\textbf{3D IN}} & \multicolumn{2}{c|}{\textbf{(2+1)D XN}} & \multicolumn{2}{c|}{\textbf{Attention GCN}} & \multicolumn{2}{c}{\textbf{Proposed}} \\ \cline{2-19} 
    & Val & Test & Val & Test & Val & Test & Val & Test & Val & Test & Val & Test & Val & Test & Val & Test & Val & Test \\ \hline
    
    Asphalt & 99.75 & 99.72 & 99.93 & 99.90 & 98.61 & 98.58 & 100 & 99.93 & 99.57 & 99.75 & 100 & 99.93 & 99.87 & 99.87 & 99.81 & 99.87 & 99.93 & 99.90 \\
    Meadows & 99.95 & 99.97 & 100 & 100 & 99.97 & 99.96 & 100 & 100 & 99.95 & 99.96 & 100 & 100 & 100 & 99.98 & 99.95 & 100 & 100 & 100 \\
    Gravel & 97.52 & 97.23 & 98.28 & 98.00 & 93.52 & 92.66 & 100 & 99.90 & 96.00 & 96.85 & 100 & 99.80 & 100 & 99.61 & 74.85 & 77.14 & 100 & 100 \\
    Trees & 99.08 & 99.41 & 99.60 & 99.54 & 98.82 & 99.08 & 99.86 & 99.80 & 98.56 & 99.21 & 99.73 & 100 & 99.47 & 99.54 & 95.82 & 95.62 & 100 & 99.93 \\
    Painted & 100 & 100 & 100 & 99.85 & 100 & 100 & 100 & 100 & 100 & 100 & 100 & 100 & 100 & 100 & 99.70 & 99.85 & 100 & 100 \\
    Soil & 100 & 100 & 99.84 & 99.96 & 99.84 & 99.60 & 100 & 100 & 99.92 & 99.92 & 100 & 100 & 100 & 100 & 99.68 & 99.56 & 100 & 99.92 \\
    Bitumen & 99.69 & 98.79 & 100 & 99.54 & 97.89 & 98.94 & 100 & 99.84 & 98.19 & 98.64 & 100 & 100 & 99.39 & 99.24 & 96.09 & 95.63 & 100 & 100 \\
    Bricks & 98.80 & 98.09 & 99.45 & 99.18 & 95.00 & 95.98 & 99.67 & 99.56 & 97.82 & 98.53 & 99.56 & 99.40 & 99.78 & 99.56 & 99.13 & 98.53 & 97.06 & 97.22 \\
    Shadows & 100 & 99.78 & 100 & 99.57 & 99.57 & 98.94 & 100 & 99.57 & 98.73 & 99.36 & 100 & 100 & 100 & 100 & 100 & 100 & 100 & 99.78 \\ \hline 

    \textbf{Train (S)} & \multicolumn{2}{c|}{82.92} & \multicolumn{2}{c|}{218.22} & \multicolumn{2}{c|}{143.70} & \multicolumn{2}{c|}{505.51} & \multicolumn{2}{c|}{143.19} & \multicolumn{2}{c|}{1103.46} & \multicolumn{2}{c|}{1886.01} & \multicolumn{2}{c|}{208.12} & \multicolumn{2}{c}{327.85} \\
    \textbf{Time (S)} & 1.27 & 1.79 & 1.49 & 2.48 & 1.43 & 2.88 & 5.46 & 5.23 & 1.68 & 2.81 & 4.72 & 8.42 & 13.42 & 41.22 & 3.13 & 5.37 & 3.49 & 10.50 \\
    \textbf{Kappa} & 99.53 & 99.42 & 99.75 & 99.67 & 98.48 & 98.53 & 99.95 & 99.89 & 99.12 & 99.40 & 99.93 & 99.91 & 99.88 & 99.82 & 97.58 & 97.65 & 99.65 & 99.64 \\
    \textbf{OA} & 99.64 & 99.57 & 99.81 & 99.75 & 98.85 & 98.89 & 99.96 & 99.92 & 99.33 & 99.55 & 99.94 & 99.92 & 99.91 & 99.86 & 98.18 & 98.22 & 99.74 & 99.72 \\
    \textbf{AA} & 99.42 & 99.22 & 99.68 & 99.51 & 98.14 & 98.20 & 99.95 & 99.85 & 98.75 & 99.14 & 99.92 & 99.91 & 99.83 & 99.76 & 96.12 & 96.25 & 99.67 & 99.64 \\ \hline 
    \end{tabular}}
    \label{Tab4}
\end{table*}
\begin{figure*}[!hbt]
    \centering
	\begin{subfigure}{0.32\textwidth}
		\includegraphics[width=0.99\textwidth]{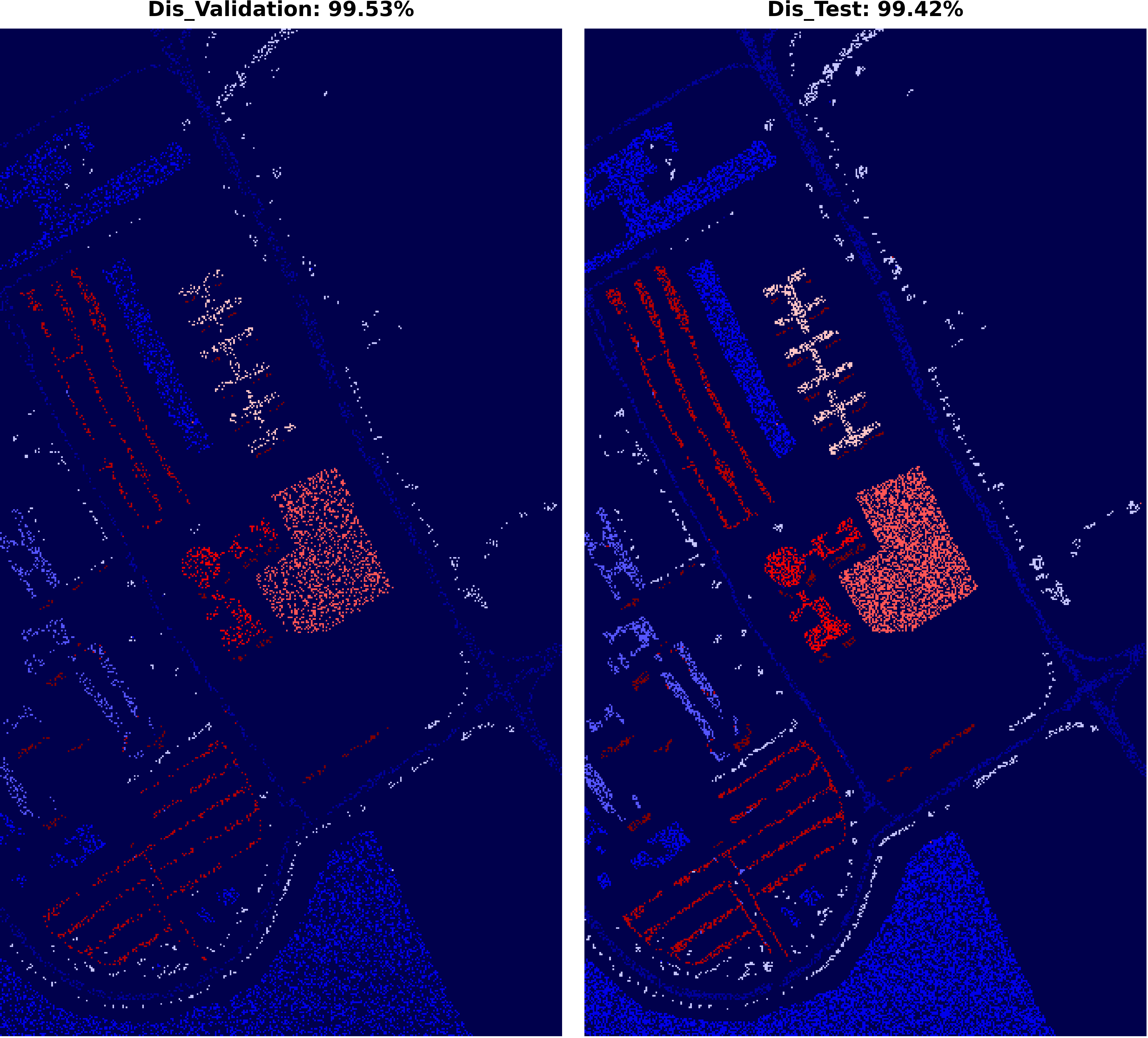}
		\caption{2D CNN} 
		\label{Fig7A}
	\end{subfigure}
	\begin{subfigure}{0.32\textwidth}
		\includegraphics[width=0.99\textwidth]{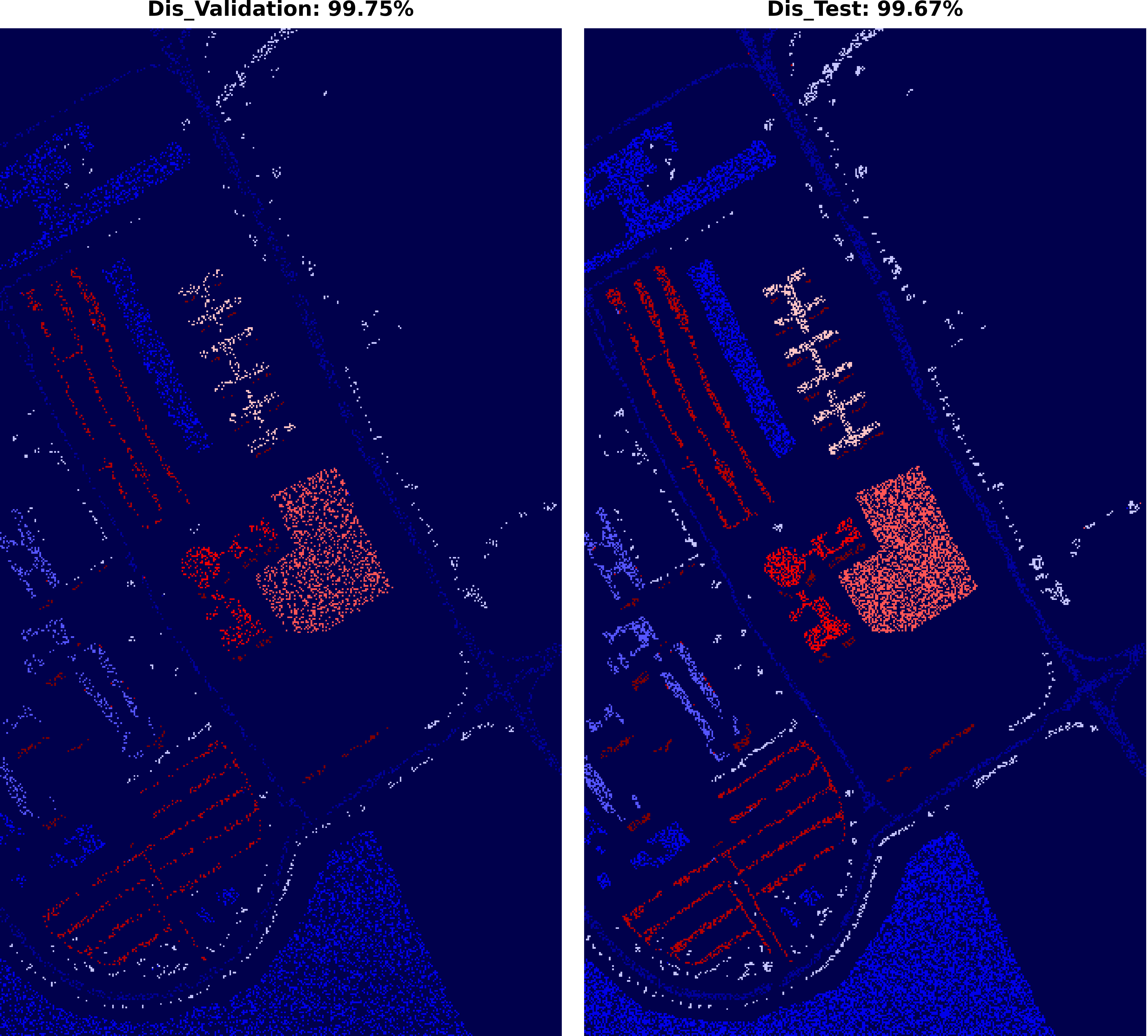}
		\caption{3D CNN}
		\label{Fig7B}
	\end{subfigure} 
	\begin{subfigure}{0.32\textwidth}
		\includegraphics[width=0.99\textwidth]{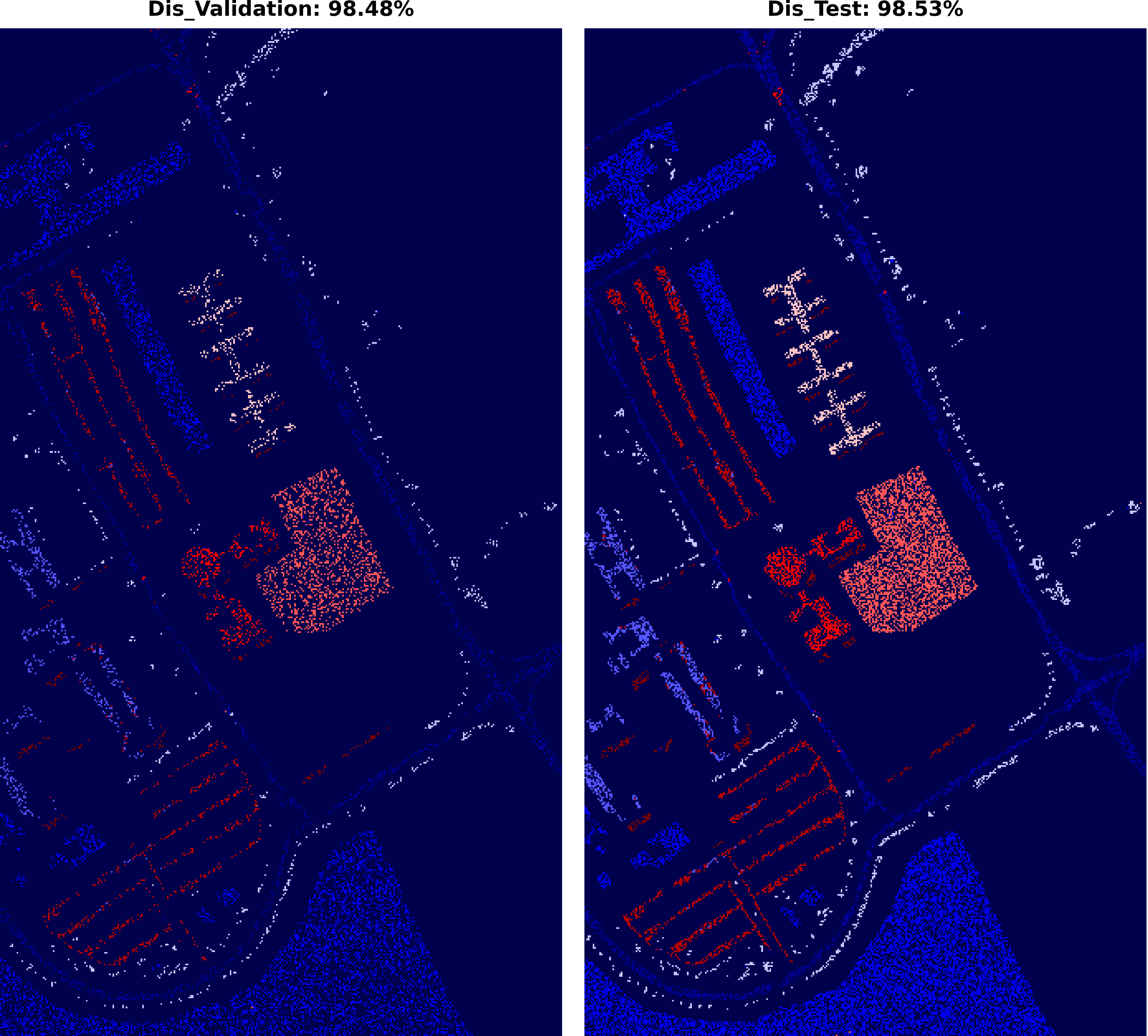}
		\caption{Hybrid CNN}
		\label{Fig7C}
	\end{subfigure} 
	\begin{subfigure}{0.32\textwidth}
		\includegraphics[width=0.99\textwidth]{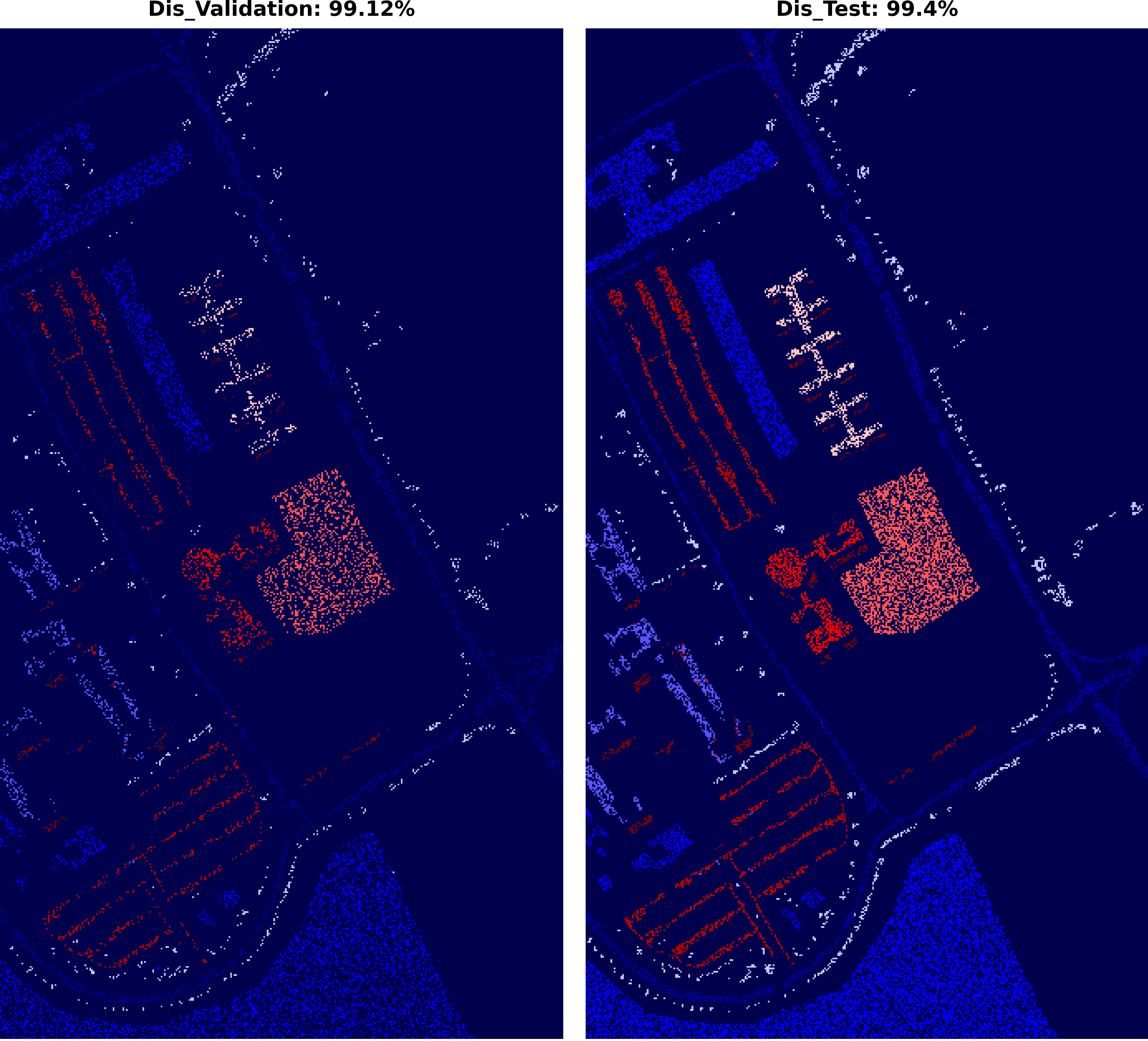}
		\caption{2D Inception Net (IN)}
		\label{Fig7D}
	\end{subfigure} 
	\begin{subfigure}{0.32\textwidth}
		\includegraphics[width=0.99\textwidth]{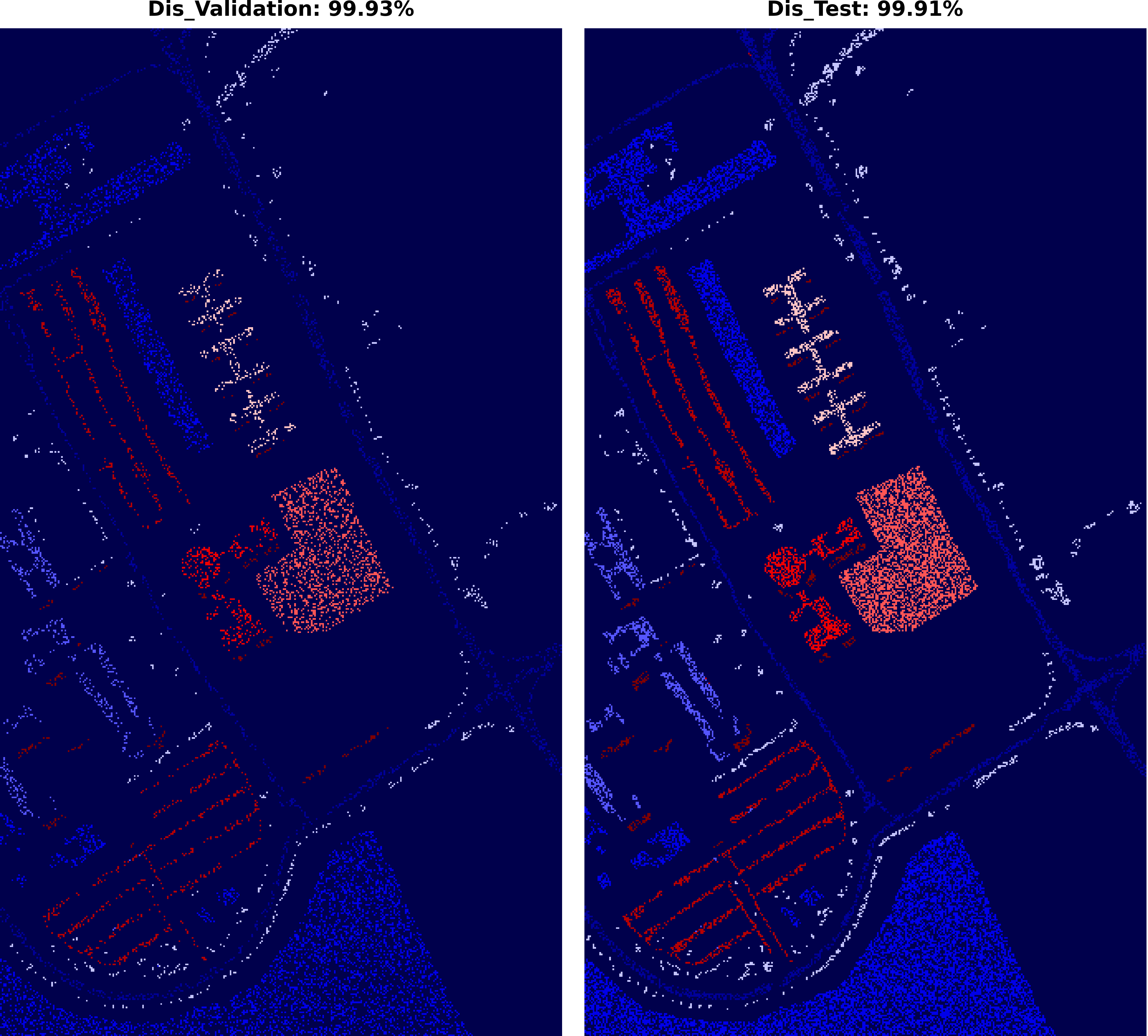}
		\caption{3D Inception Net (IN)}
		\label{Fig7E}
	\end{subfigure} 
	\begin{subfigure}{0.32\textwidth}
		\includegraphics[width=0.99\textwidth]{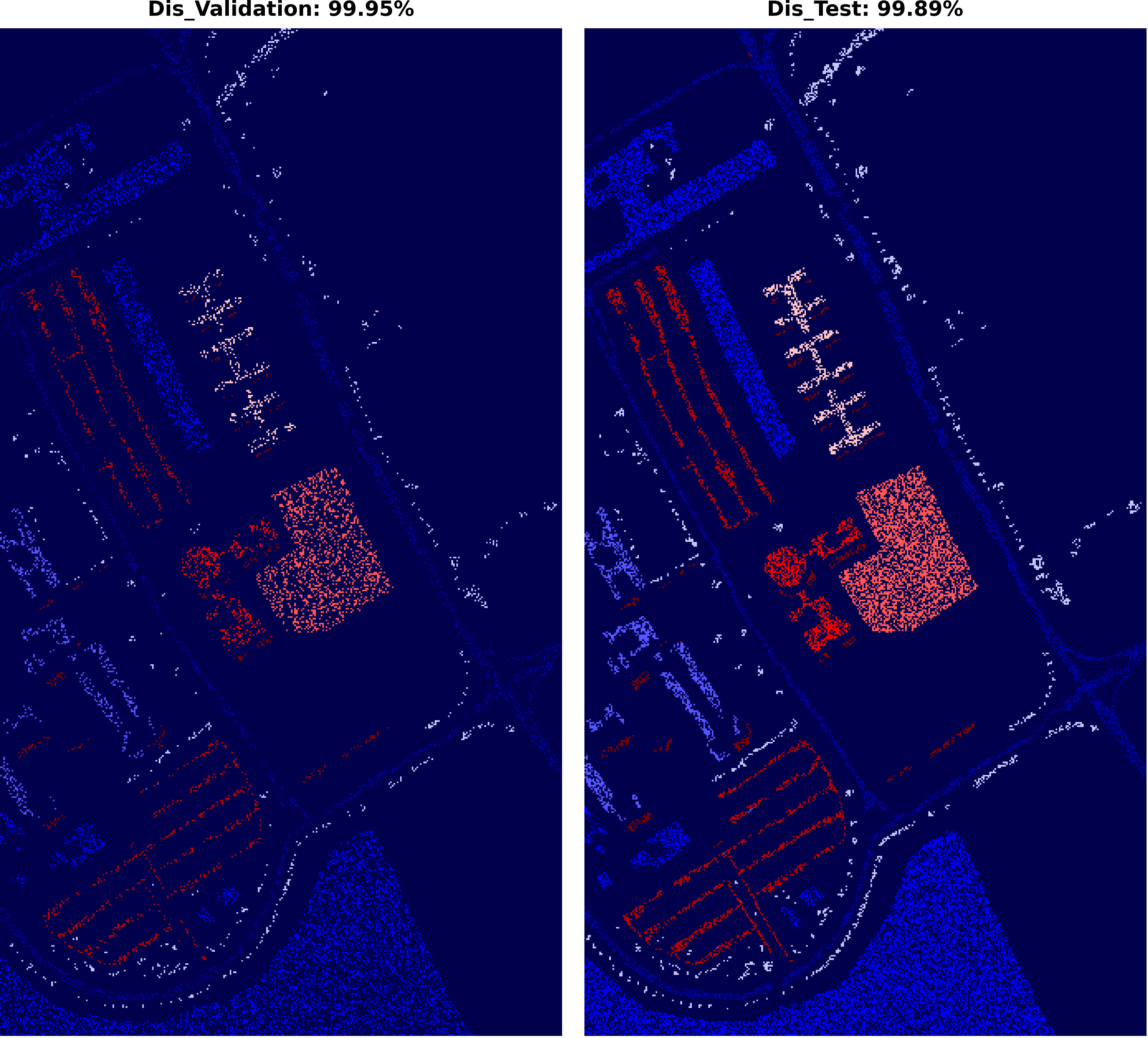}
		\caption{Hybrid Inception Net (IN)}
		\label{Fig7F}
	\end{subfigure}
    \begin{subfigure}{0.32\textwidth}
		\includegraphics[width=0.99\textwidth]{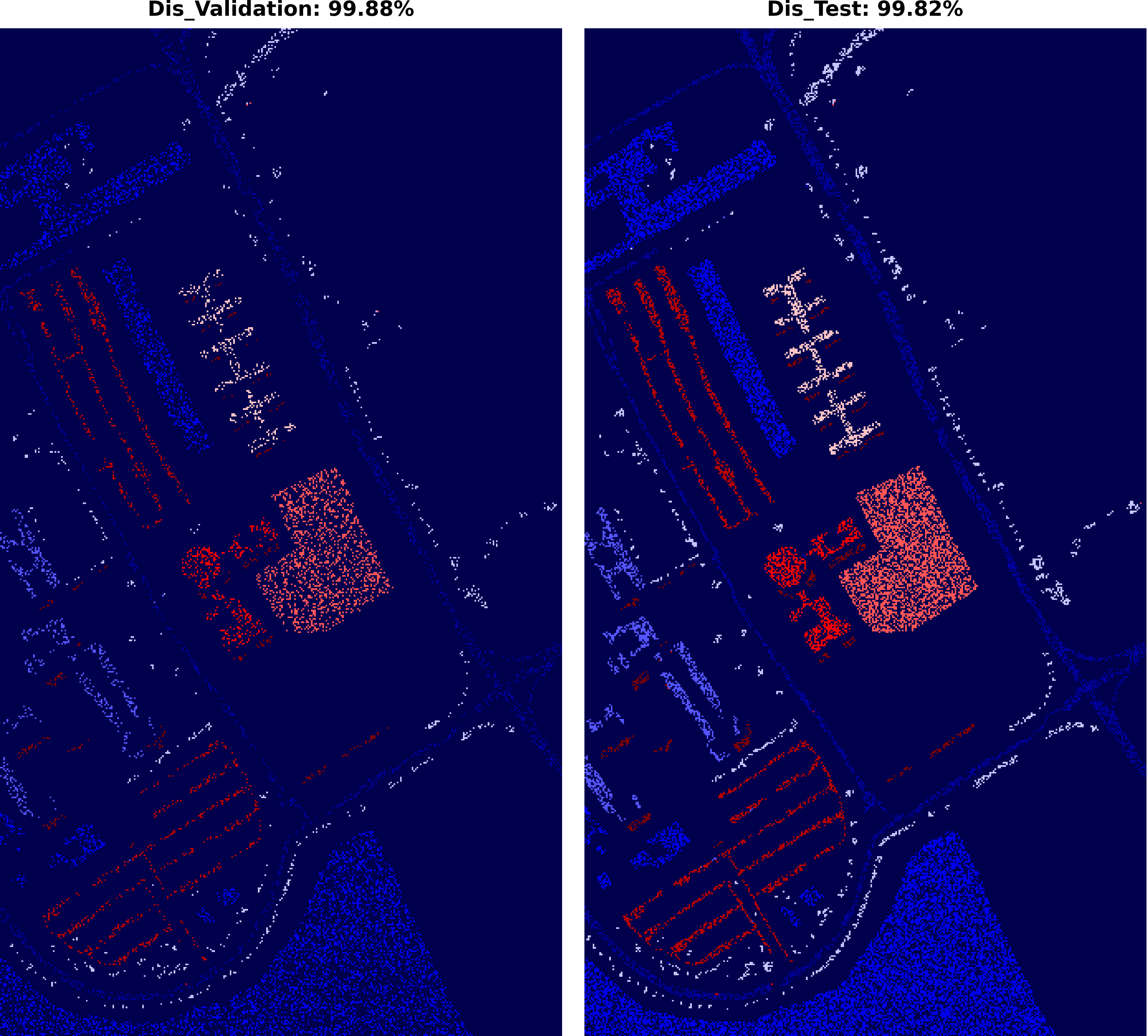}
		\caption{(2+1)D Extreme Exception Net (XN)}
		\label{Fig7G}
	\end{subfigure} 
	\begin{subfigure}{0.32\textwidth}
		\includegraphics[width=0.99\textwidth]{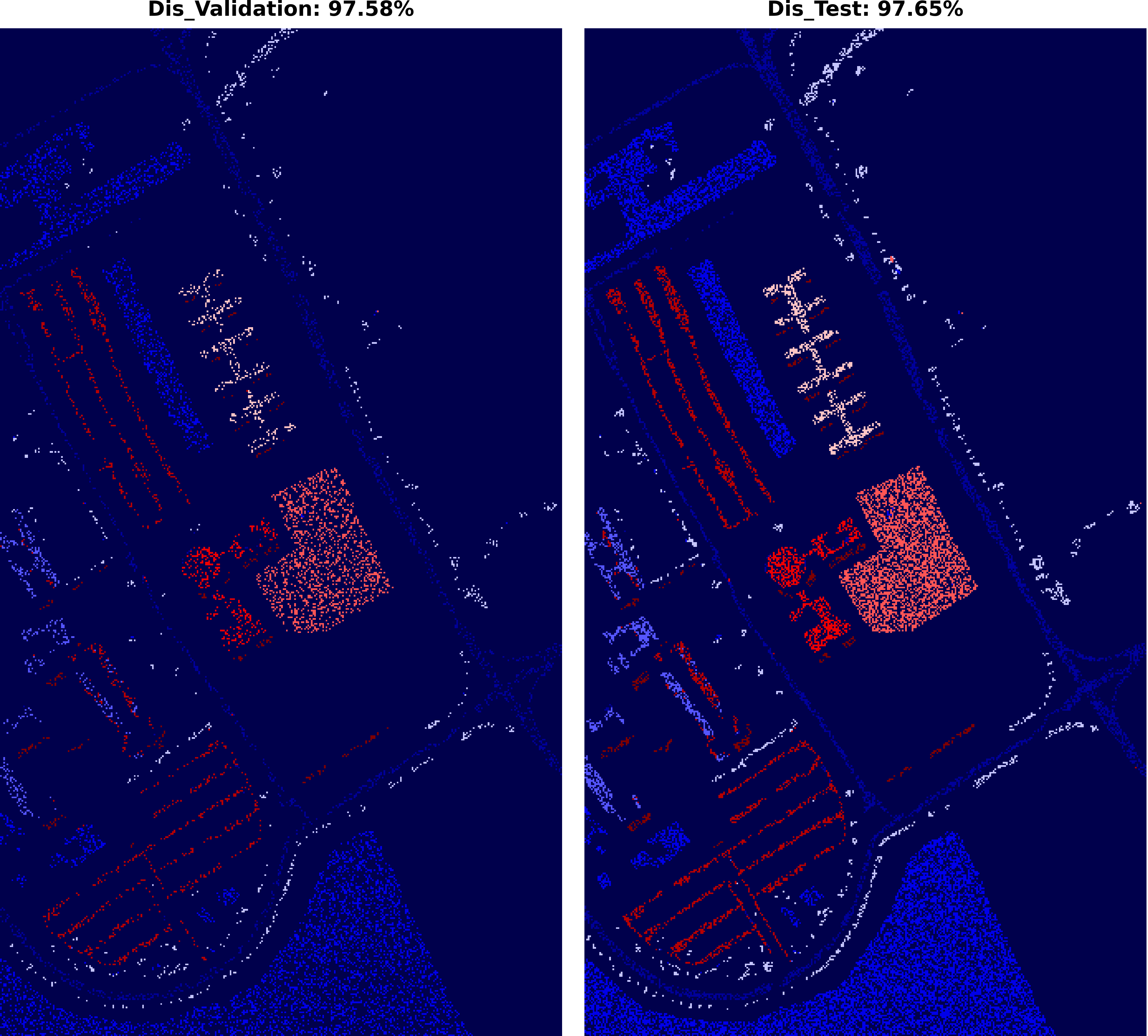}
		\caption{Attention GCN}
		\label{Fig7H}
	\end{subfigure} 
	\begin{subfigure}{0.32\textwidth}
		\includegraphics[width=0.99\textwidth]{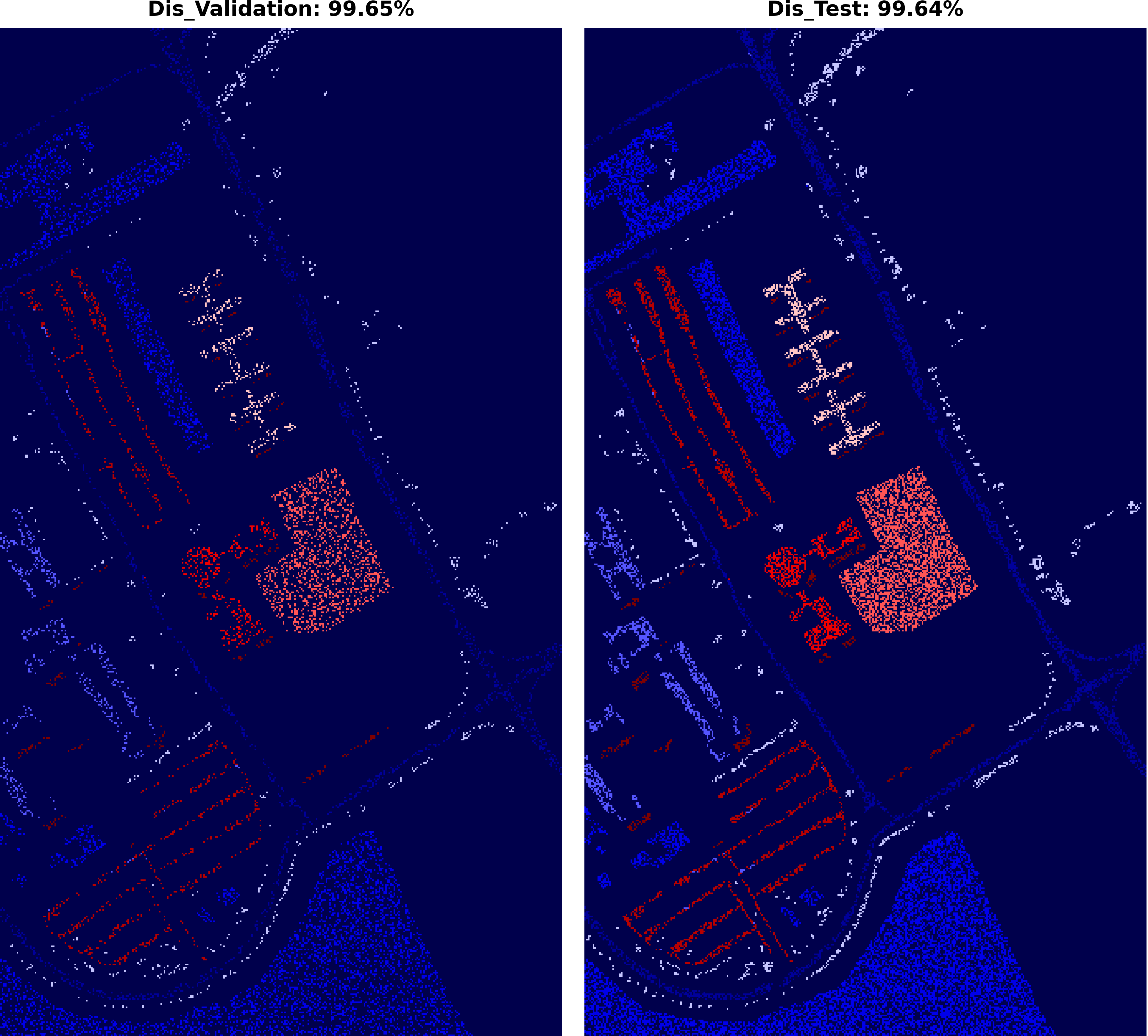}
		\caption{Proposed}
		\label{Fig7I}
	\end{subfigure} 
\caption{\textbf{Pavia University Dataset:} Land cover maps for disjoint validation and test set are provided. Comprehensive class-wise results can be found in Table \ref{Tab4}.}
\label{Fig7}
\end{figure*}

\begin{table*}[!hbt]
    \centering
    \caption{\textbf{University of Houston:} Per-Class Classification results of proposed method along with the comparative methods. Class-wise ground truth maps are presented in Figure \ref{Fig8}.}
    \resizebox{\textwidth}{!}{\begin{tabular}{c|cc|cc|cc|cc|cc|cc|cc|cc|cc} \hline 
    \multirow{2}{*}{\textbf{Class}} & \multicolumn{2}{c|}{\textbf{2D CNN}} & \multicolumn{2}{c|}{\textbf{3D CNN}} & \multicolumn{2}{c|}{\textbf{Hybrid CNN}} & \multicolumn{2}{c|}{\textbf{Hybrid IN}} & \multicolumn{2}{c|}{\textbf{2D IN}} & \multicolumn{2}{c|}{\textbf{3D IN}} & \multicolumn{2}{c|}{\textbf{(2+1)D XN}} & \multicolumn{2}{c|}{\textbf{Attention GCN}} & \multicolumn{2}{c}{\textbf{Proposed}} \\ \cline{2-19} 
    
    & Val & Test & Val & Test & Val & Test & Val & Test & Val & Test & Val & Test & Val & Test & Val & Test & Val & Test \\ \hline
    
    Healthy grass & 98.40 & 98.72 & 100 & 99.20 & 99.36 & 99.04 & 100 & 100 & 100 & 99.68 & 100 & 100 & 100 & 100 & --- & --- & 100 & 100 \\
    Stressed grass & 99.36 & 99.68 & 99.68 & 100 & 99.36 & 99.68 & 99.68 & 100 & 99.68 & 100 & 99.68 & 100 & 99.68 & 100 & --- & --- & 99.68 & 100 \\
    Synthetic grass & 100 & 100 & 100 & 100 & 100 & 99.71 & 100 & 100 & 100 & 100 & 100 & 100 & 100 & 100 & --- & --- & 100 & 100 \\
    Trees & 98.39 & 99.51 & 99.03 & 98.71 & 99.67 & 99.83 & 99.35 & 100 & 99.03 & 99.67 & 99.67 & 100 & 99.35 & 99.67 & --- & --- & 99.03 & 99.03 \\
    Soil & 100 & 100 & 100 & 99.83 & 100 & 100 & 100 & 100 & 100 & 99.83 & 100 & 100 & 100 & 100 & --- & --- & 100 & 100 \\
    Water & 98.76 & 97.54 & 100 & 99.38 & 98.76 & 100 & 100 & 100 & 100 & 99.38 & 100 & 100 & 100 & 100 & --- & --- & 100 & 100 \\
    Residential & 96.84 & 96.37 & 99.05 & 98.73 & 97.79 & 97.94 & 100 & 99.84 & 100 & 98.73 & 100 & 99.84 & 99.68 & 99.36 & 100 & 100 & 100 & 100 \\
    Commercial & 97.74 & 95.98 & 99.35 & 98.39 & 99.67 & 99.03 & 99.67 & 98.87 & 98.07 & 96.78 & 100 & 100 & 99.67 & 99.03 & --- & --- & 99.67 & 98.55 \\
    Road & 93.29 & 94.08 & 99.68 & 99.84 & 99.68 & 100 & 100 & 100 & 99.36 & 99.68 & 99.68 & 100 & 100 & 100 & --- & --- & 100 & 99.84 \\
    Highway & 99.67 & 98.37 & 99.67 & 99.67 & 100 & 100 & 100 & 100 & 99.67 & 99.67 & 100 & 100 & 99.02 & 99.34 & --- & --- & 100 & 100 \\
    Railway & 98.70 & 99.02 & 100 & 100 & 99.35 & 99.67 & 99.67 & 100 & 98.70 & 99.51 & 99.67 & 100 & 100 & 100 & --- & --- & 100 & 100 \\
    Parking Lot 1 & 98.37 & 98.86 & 100 & 99.83 & 99.67 & 99.67 & 99.67 & 100 & 99.67 & 99.67 & 100 & 100 & 100 & 100 & --- & --- & 98.05 & 97.40 \\
    Parking Lot 2 & 94.01 & 95.31 & 97.43 & 100 & 87.17 & 89.78 & 100 & 100 & 88.03 & 94.89 & 100 & 100 & 100 & 100 & --- & --- & 96.58 & 97.87 \\
    Tennis Court & 100 & 99.06 & 99.06 & 98.59 & 100 & 100 & 100 & 100 & 99.06 & 99.53 & 100 & 100 & 99.06 & 100 & --- & --- & 100 & 100 \\
    Running Track & 100 & 100 & 100 & 100 & 100 & 100 & 100 & 100 & 100 & 100 & 100 & 100 & 100 & 100 & --- & --- & 100 & 100 \\ \hline 

    \textbf{Train (S)} & \multicolumn{2}{c|}{41.75} & \multicolumn{2}{c|}{84.82} & \multicolumn{2}{c|}{82.59} & \multicolumn{2}{c|}{178.18} & \multicolumn{2}{c|}{41.88} & \multicolumn{2}{c|}{443.46} & \multicolumn{2}{c|}{745.89} & \multicolumn{2}{c|}{61.70} & \multicolumn{2}{c}{149.73} \\
    \textbf{Time (S)} & 0.71 & 0.73 & 0.64 & 1.41 & 0.51 & 1.38 & 1.13 & 1.86 & 1.43 & 0.91 & 2.82 & 3.26 & 6.92 & 9.00 & 1.10 & 1.14 & 2.35 & 2.07 \\
    \textbf{Kappa} & 98.04 & 98.02 & 99.57 & 99.42 & 99.05 & 99.18 & 99.83 & 99.88 & 99.05 & 99.19 & 99.88 & 99.99 & 99.74 & 99.77 & --- & --- & 99.57 & 99.47 \\
    \textbf{OA} & 98.19 & 98.16 & 99.60 & 99.47 & 99.11 & 99.24 & 99.83 & 99.89 & 99.11 & 99.26 & 99.89 & 99.99 & 99.76 & 99.79 & 8.44 & 8.43 & 99.60 & 99.51 \\
    \textbf{AA} & 98.24 & 98.17 & 99.53 & 99.48 & 98.70 & 98.96 & 99.87 & 99.91 & 98.75 & 99.14 & 99.91 & 99.99 & 99.77 & 99.83 & 6.67 & 6.67 & 99.53 & 99.51 \\ \hline 
    \end{tabular}}
    \label{Tab5}
\end{table*}
\begin{figure*}[!hbt]
    \centering
	\begin{subfigure}{0.32\textwidth}
		\includegraphics[width=0.99\textwidth]{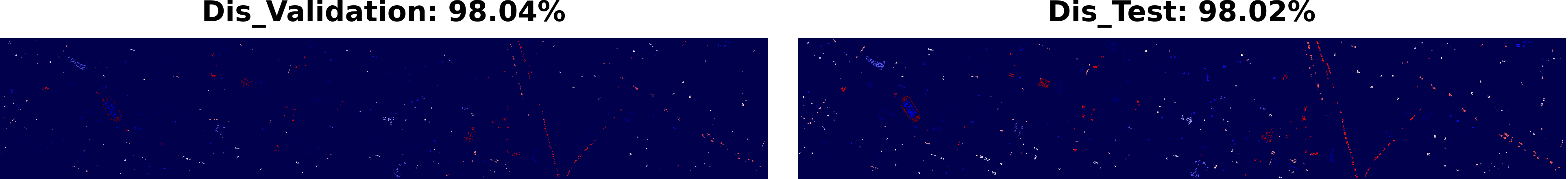}
		\caption{2D CNN} 
		\label{Fig8A}
	\end{subfigure}
	\begin{subfigure}{0.32\textwidth}
		\includegraphics[width=0.99\textwidth]{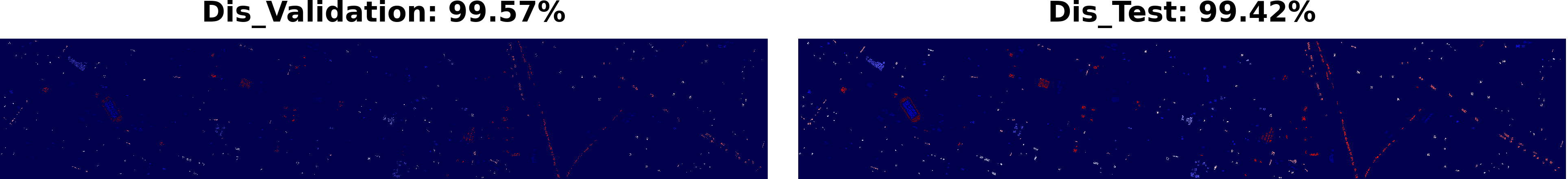}
		\caption{3D CNN}
		\label{Fig8B}
	\end{subfigure} 
	\begin{subfigure}{0.32\textwidth}
		\includegraphics[width=0.99\textwidth]{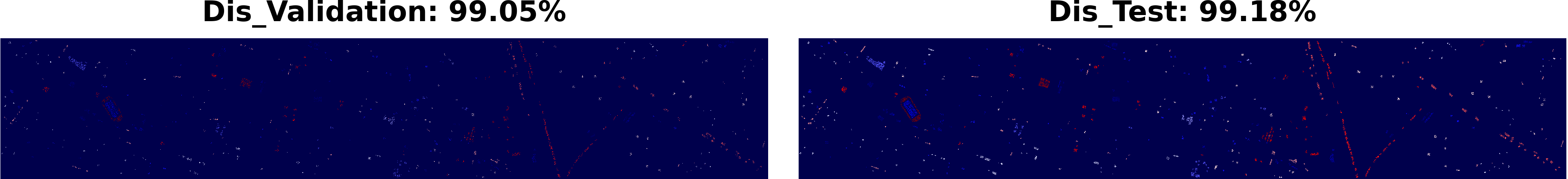}
		\caption{Hybrid CNN}
		\label{Fig8C}
	\end{subfigure} 
	\begin{subfigure}{0.32\textwidth}
		\includegraphics[width=0.99\textwidth]{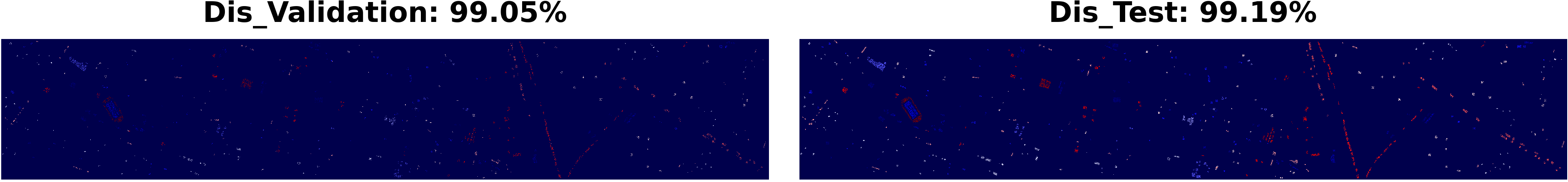}
		\caption{2D Inception Net (IN)}
		\label{Fig8D}
	\end{subfigure} 
	\begin{subfigure}{0.32\textwidth}
		\includegraphics[width=0.99\textwidth]{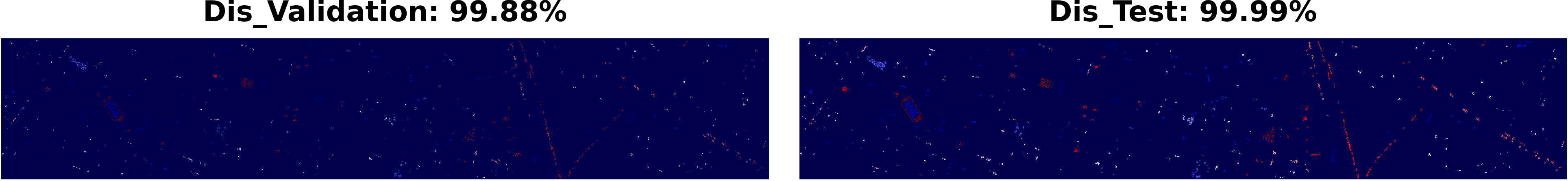}
		\caption{3D Inception Net (IN)}
		\label{Fig8E}
	\end{subfigure} 
	\begin{subfigure}{0.32\textwidth}
		\includegraphics[width=0.99\textwidth]{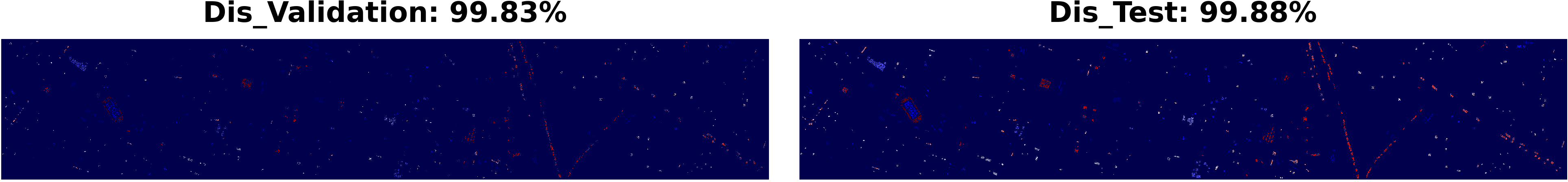}
		\caption{Hybrid Inception Net (IN)}
		\label{Fig8F}
	\end{subfigure}
    \begin{subfigure}{0.32\textwidth}
		\includegraphics[width=0.99\textwidth]{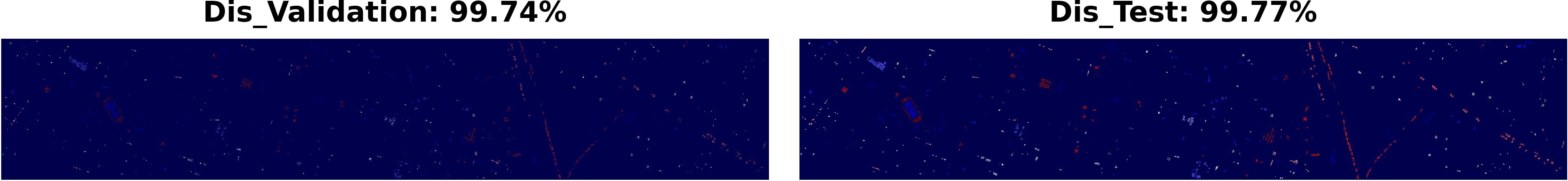}
		\caption{(2+1)D Extreme Exception Net (XN)}
		\label{Fig8G}
	\end{subfigure} 
	\begin{subfigure}{0.32\textwidth}
		\includegraphics[width=0.99\textwidth]{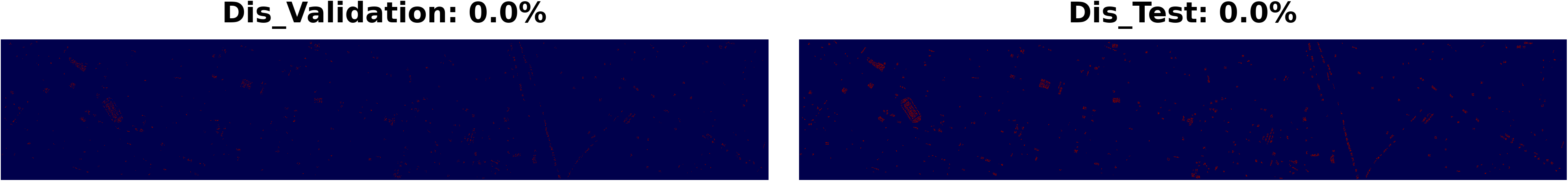}
		\caption{Attention GCN}
		\label{Fig8H}
	\end{subfigure} 
	\begin{subfigure}{0.32\textwidth}
		\includegraphics[width=0.99\textwidth]{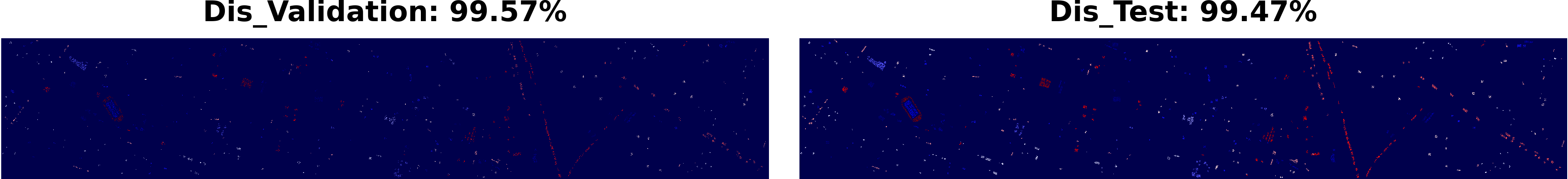}
		\caption{Proposed}
		\label{Fig8I}
	\end{subfigure} 
\caption{\textbf{University of Houston Dataset:} Land cover maps for disjoint validation and test set are provided. Comprehensive class-wise results can be found in Table \ref{Tab5}.}
\label{Fig8}
\end{figure*}

\begin{table*}[!hbt]
    \centering
    \caption{\textbf{Salinas:} Per-Class Classification results of proposed method along with the comparative methods. Class-wise ground truth maps are presented in Figure \ref{Fig9}.}
    \resizebox{\textwidth}{!}{\begin{tabular}{c|cc|cc|cc|cc|cc|cc|cc|cc|cc} \hline 
    \multirow{2}{*}{\textbf{Class}} & \multicolumn{2}{c|}{\textbf{2D CNN}} & \multicolumn{2}{c|}{\textbf{3D CNN}} & \multicolumn{2}{c|}{\textbf{Hybrid CNN}} & \multicolumn{2}{c|}{\textbf{Hybrid IN}} & \multicolumn{2}{c|}{\textbf{2D IN}} & \multicolumn{2}{c|}{\textbf{3D IN}} & \multicolumn{2}{c|}{\textbf{(2+1)D XN}} & \multicolumn{2}{c|}{\textbf{Attention GCN}} & \multicolumn{2}{c}{\textbf{Proposed}} \\ \cline{2-19} 
    
    & Val & Test & Val & Test & Val & Test & Val & Test & Val & Test & Val & Test & Val & Test & Val & Test & Val & Test \\ \hline
    
    Weeds 1 & 100 & 100 & 99.80 & 100 & 100 & 100 & 100 & 100 & 100 & 100 & 100 & 100 & 100 & 100 & 100 & 100 & 100 & 100 \\
    Weeds 2 & 100 & 100 & 100 & 100 & 100 & 100 & 100 & 100 & 100 & 99.94 & 100 & 100 & 100 & 100 & 100 & 99.83 & 100 & 100 \\
    Fallow & 100 & 100 & 100 & 100 & 100 & 100 & 100 & 100 & 100 & 100 & 100 & 100 & 100 & 100 & 100 & 100 & 100 & 100 \\
    Fallow rough plow & 99.71 & 99.56 & 100 & 99.71 & 99.14 & 99.85 & 100 & 99.85 & 99.71 & 99.42 & 100 & 100 & 100 & 100 & 99.42 & 99.42 & 100 & 100 \\
    Fallow smooth & 99.70 & 99.25 & 100 & 100 & 99.25 & 99.02 & 100 & 100 & 98.35 & 98.58 & 99.40 & 99.77 & 100 & 100 & 99.25 & 99.85 & 98.95 & 99.62 \\
    Stubble & 100 & 100 & 100 & 100 & 100 & 100 & 100 & 100 & 100 & 100 & 100 & 100 & 100 & 100 & 100 & 99.89 & 100 & 100 \\
    Celery & 100 & 100 & 100 & 100 & 100 & 100 & 100 & 100 & 100 & 100 & 100 & 100 & 100 & 100 & 99.88 & 99.94 & 100 & 100 \\
    Grapes untrained & 99.89 & 99.76 & 99.92 & 99.92 & 98.47 & 98.33 & 99.96 & 99.92 & 99.57 & 99.66 & 100 & 99.96 & 99.96 & 99.89 & 97.44 & 97.24 & 100 & 100 \\
    Soil vinyard develop & 100 & 100 & 100 & 100 & 100 & 100 & 100 & 100 & 100 & 100 & 100 & 100 & 100 & 100 & 100 & 100 & 100 & 100 \\
    Corn Weeds & 100 & 100 & 100 & 99.93 & 99.87 & 100 & 100 & 100 & 100 & 100 & 100 & 100 & 99.87 & 99.93 & 99.87 & 99.93 & 100 & 100 \\
    Lettuce 4wk & 100 & 100 & 100 & 100 & 100 & 100 & 100 & 100 & 100 & 100 & 100 & 100 & 100 & 100 & 98.87 & 100 & 100 & 100 \\
    Lettuce 5wk & 100 & 100 & 100 & 100 & 100 & 100 & 100 & 100 & 100 & 100 & 100 & 100 & 100 & 100 & 100 & 99.89 & 100 & 100 \\
    Lettuce 6wk & 100 & 100 & 100 & 100 & 100 & 100 & 100 & 100 & 100 & 100 & 100 & 100 & 100 & 100 & 100 & 99.78 & 100 & 100 \\
    Lettuce 7wk & 100 & 99.62 & 99.62 & 99.62 & 100 & 99.81 & 100 & 99.81 & 100 & 100 & 100 & 99.81 & 100 & 100 & 99.25 & 99.62 & 100 & 100 \\
    Vinyard untrained & 98.89 & 98.37 & 99.50 & 99.20 & 97.41 & 97.02 & 99.83 & 99.53 & 97.96 & 97.55 & 99.61 & 99.44 & 99.72 & 99.47 & 95.10 & 94.22 & 99.44 & 99.36 \\
    Vinyard trellis & 100 & 100 & 100 & 100 & 100 & 100 & 100 & 100 & 100 & 100 & 100 & 100 & 100 & 100 & 100 & 99.77 & 100 & 100 \\ \hline 

    \textbf{Train (S)} & \multicolumn{2}{c|}{84.32} & \multicolumn{2}{c|}{287.88} & \multicolumn{2}{c|}{202.75} & \multicolumn{2}{c|}{639.97} & \multicolumn{2}{c|}{130.13} & \multicolumn{2}{c|}{1496.13} & \multicolumn{2}{c|}{2333.72} & \multicolumn{2}{c|}{201.07} & \multicolumn{2}{c}{325.29} \\
    \textbf{Time (S)} & 1.51 & 2.93 & 2.80 & 3.62 & 1.53 & 2.44 & 5.48 & 10.86 & 2.84 & 2.98 & 5.91 & 11.57 & 15.87 & 29.31 & 2.86 & 5.85 & 2.75 & 8.75 \\
    \textbf{Kappa} & 99.79 & 99.64 & 99.89 & 99.84 & 99.19 & 99.11 & 99.97 & 99.91 & 99.50 & 99.46 & 99.91 & 99.89 & 99.94 & 99.89 & 98.56 & 98.42 & 99.86 & 99.88 \\
    \textbf{OA} & 99.81 & 99.68 & 99.90 & 99.86 & 99.27 & 99.20 & 99.97 & 99.92 & 99.55 & 99.51 & 99.92 & 99.90 & 99.95 & 99.90 & 98.71 & 98.58 & 99.87 & 99.90 \\
    \textbf{AA} & 99.89 & 99.79 & 99.92 & 99.90 & 99.63 & 99.63 & 99.99 & 99.95 & 99.72 & 99.70 & 99.94 & 99.94 & 99.97 & 99.96 & 99.32 & 99.33 & 99.90 & 99.94 \\ \hline 
    
    \end{tabular}}
    \label{Tab6}
\end{table*}
\begin{figure*}[!hbt]
    \centering
	\begin{subfigure}{0.32\textwidth}
		\includegraphics[width=0.99\textwidth]{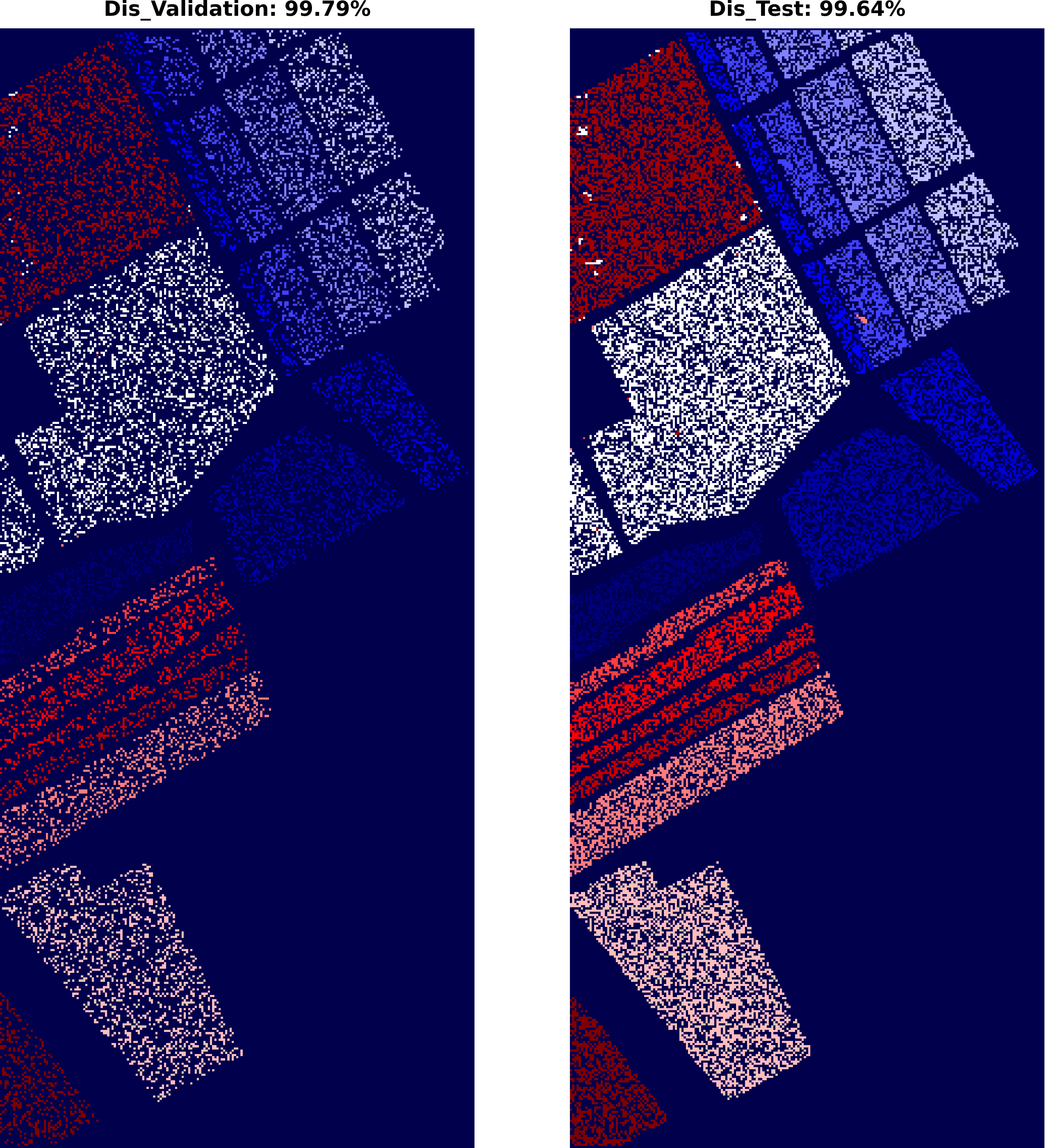}
		\caption{2D CNN} 
		\label{Fig9A}
	\end{subfigure}
	\begin{subfigure}{0.32\textwidth}
		\includegraphics[width=0.99\textwidth]{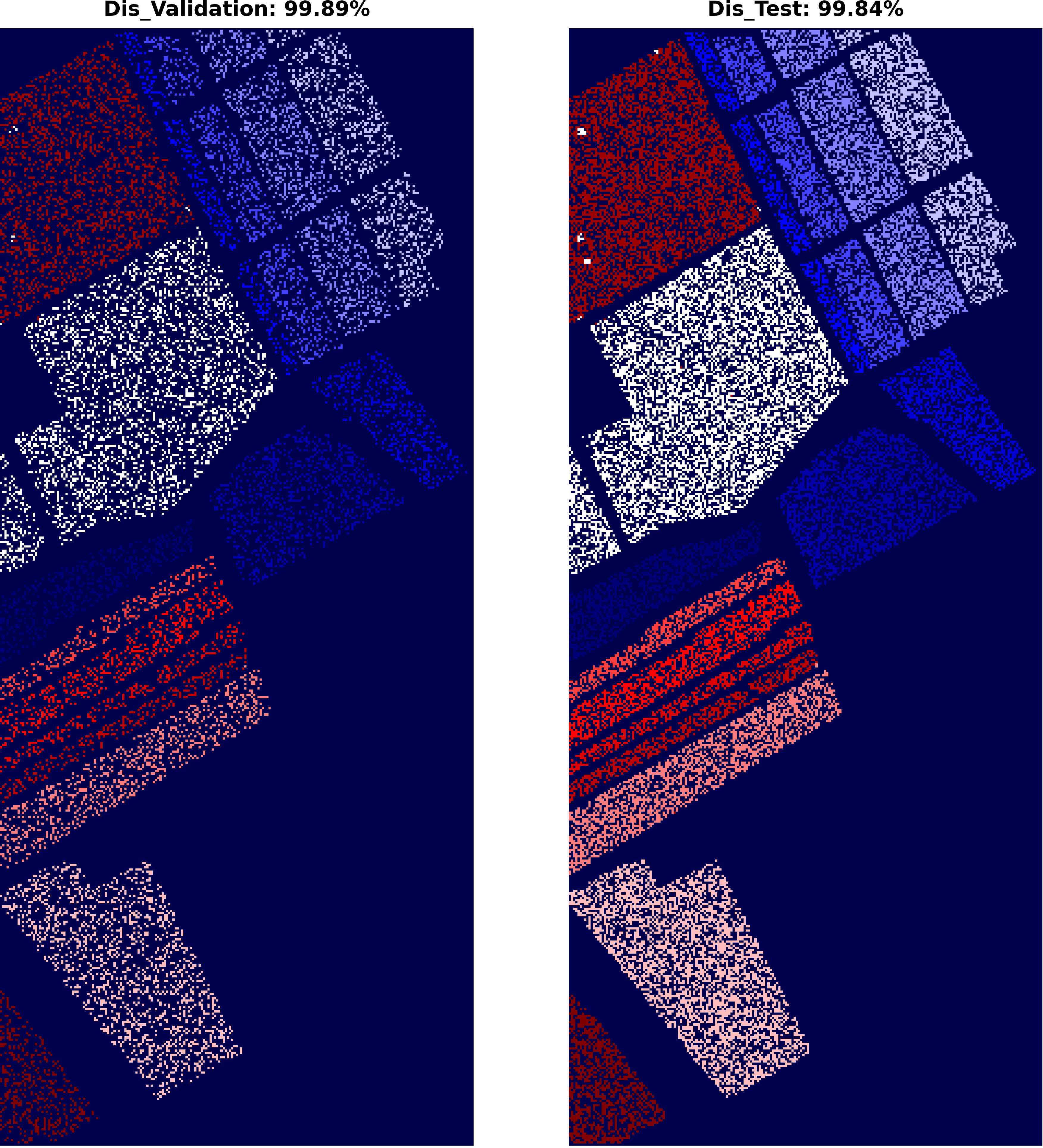}
		\caption{3D CNN}
		\label{Fig9B}
	\end{subfigure} 
	\begin{subfigure}{0.32\textwidth}
		\includegraphics[width=0.99\textwidth]{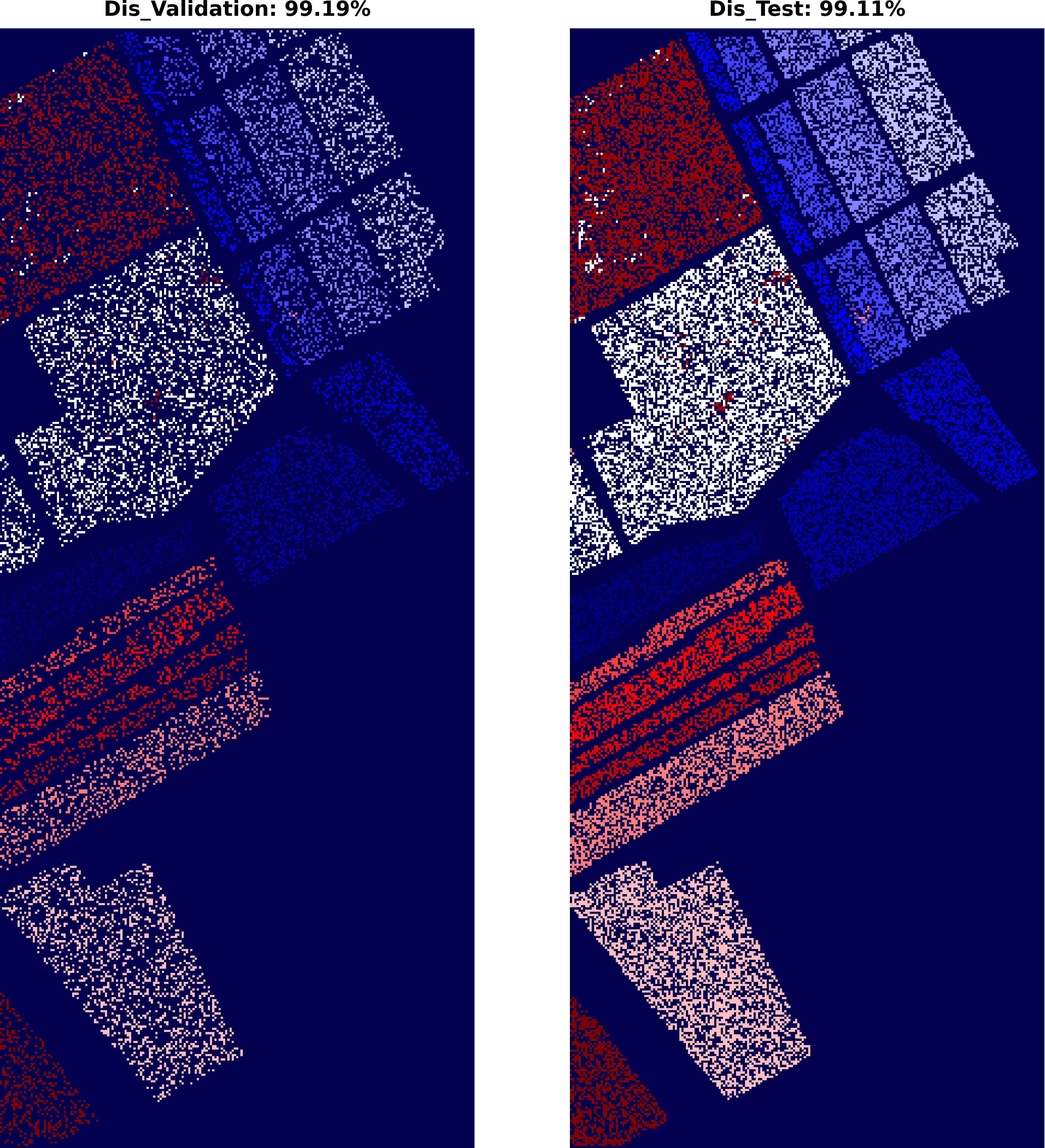}
		\caption{Hybrid CNN}
		\label{Fig9C}
	\end{subfigure} 
	\begin{subfigure}{0.32\textwidth}
		\includegraphics[width=0.99\textwidth]{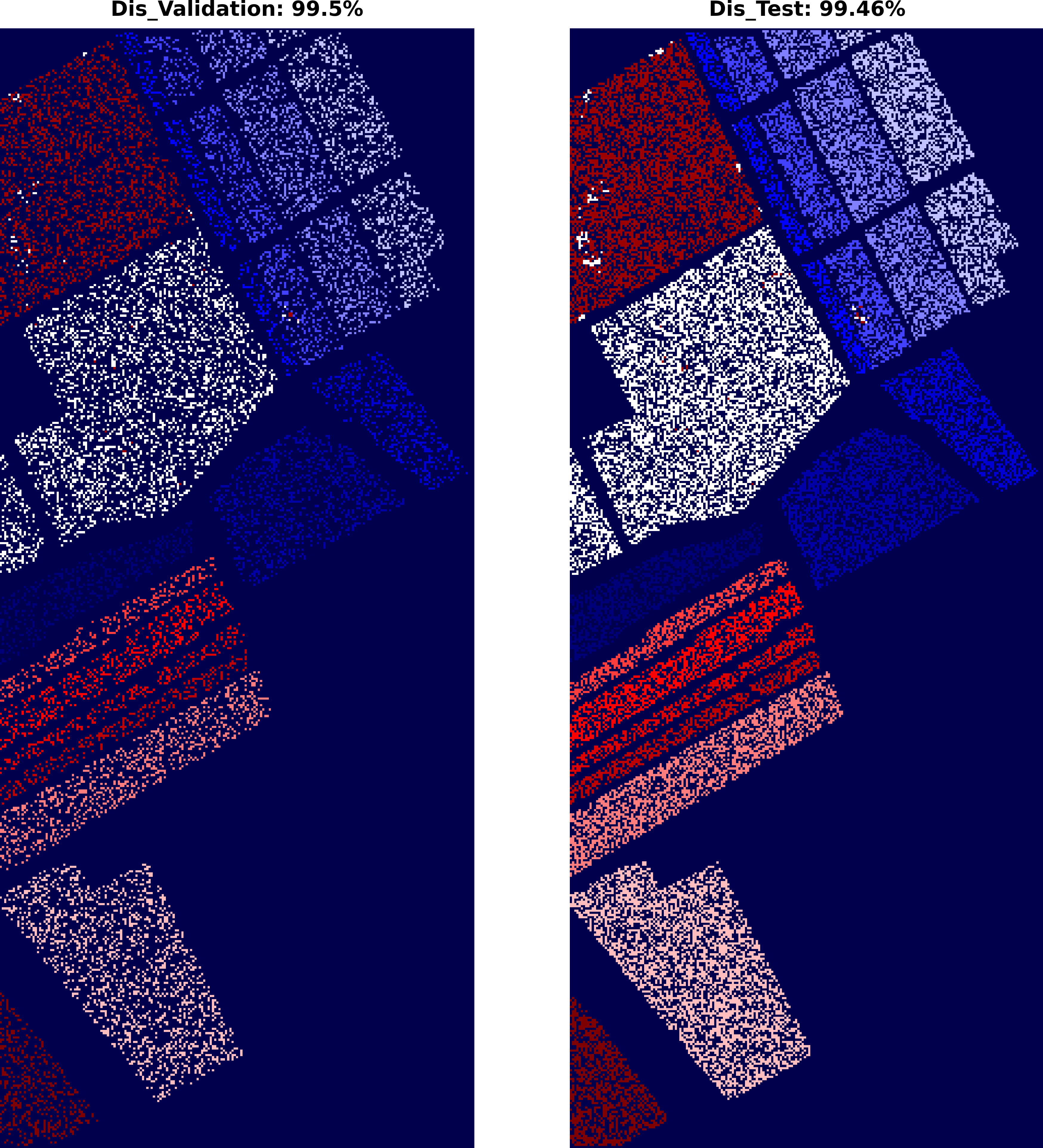}
		\caption{2D Inception Net (IN)}
		\label{Fig9D}
	\end{subfigure} 
	\begin{subfigure}{0.32\textwidth}
		\includegraphics[width=0.99\textwidth]{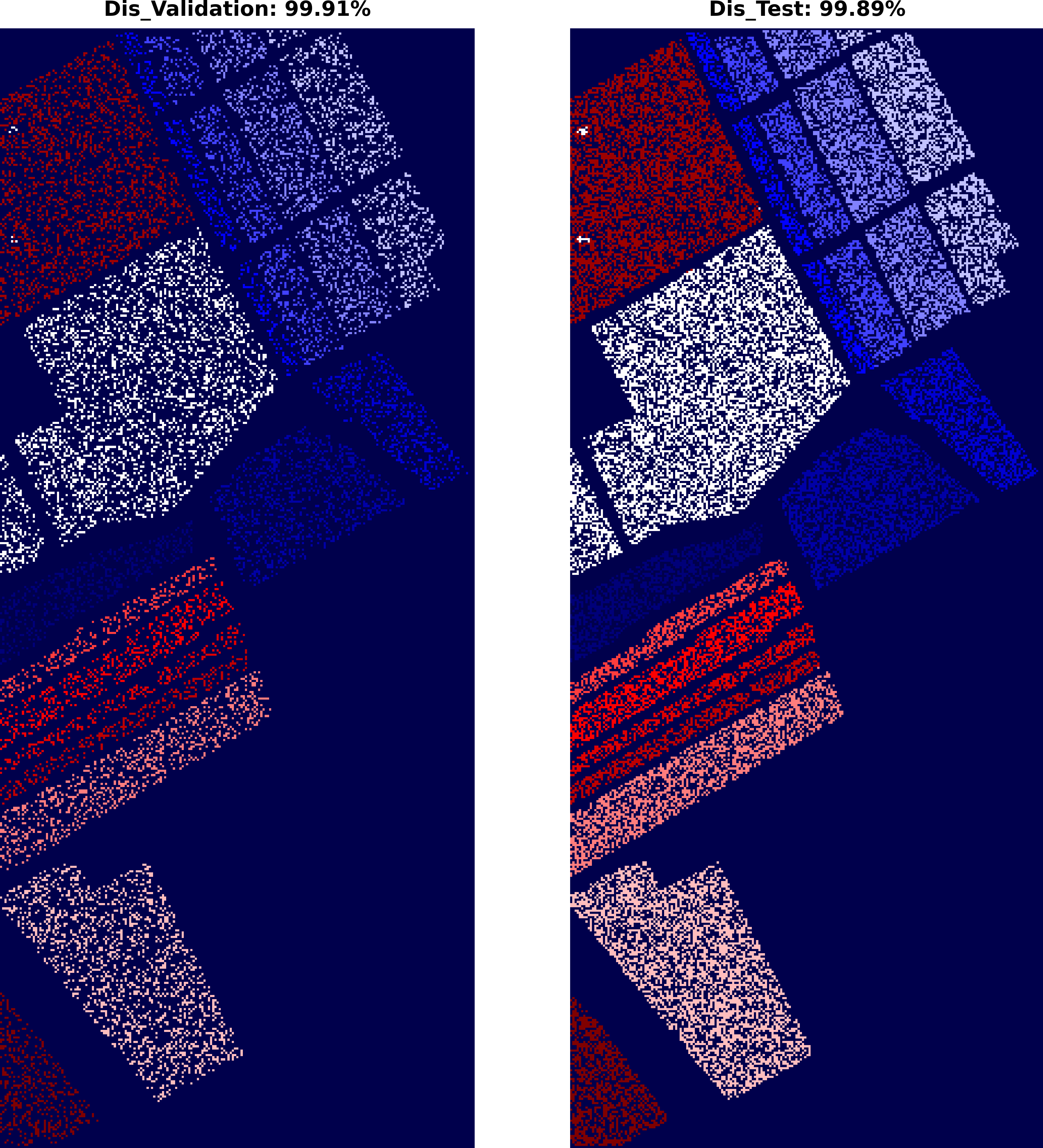}
		\caption{3D Inception Net (IN)}
		\label{Fig9E}
	\end{subfigure} 
	\begin{subfigure}{0.32\textwidth}
		\includegraphics[width=0.99\textwidth]{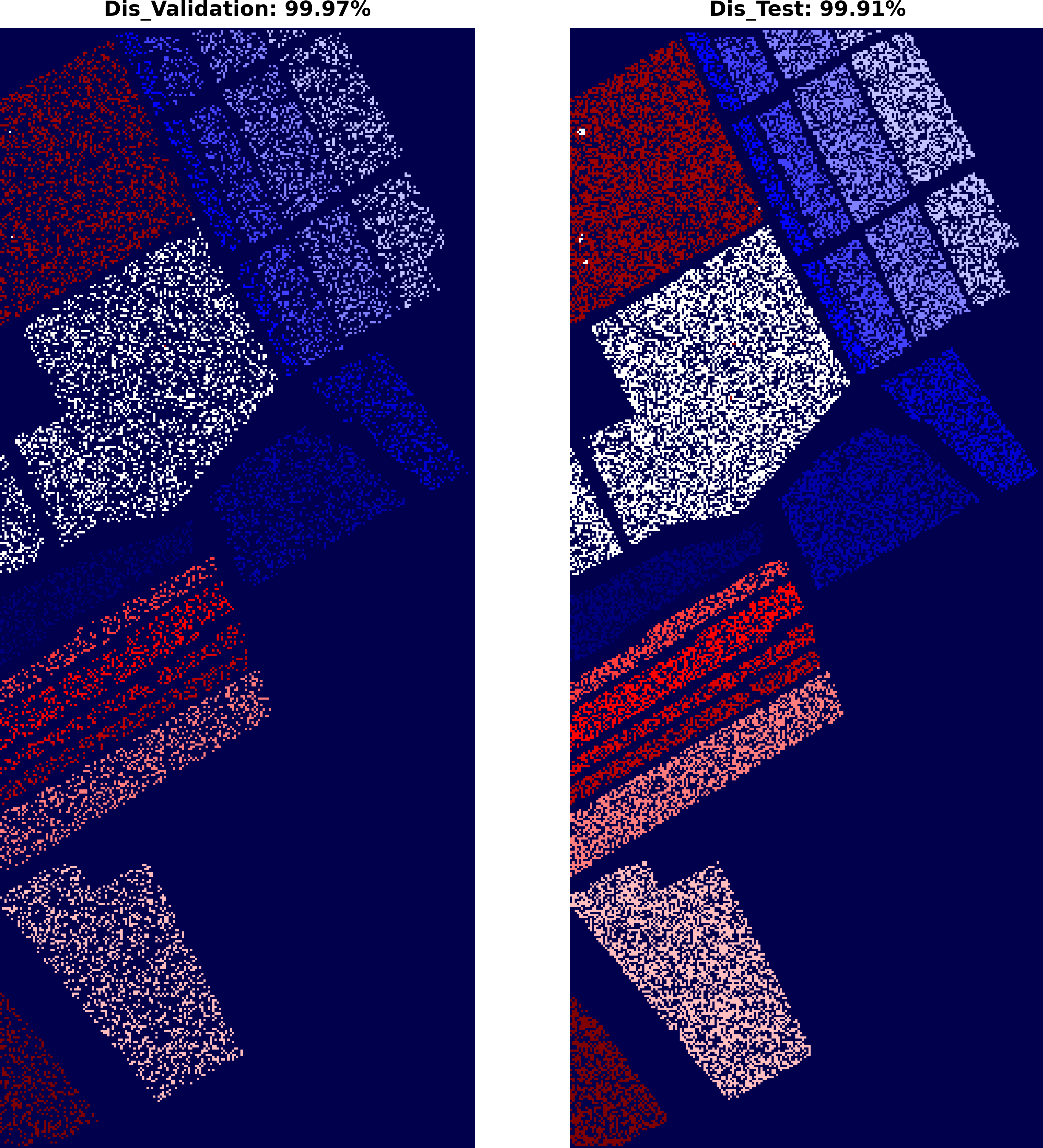}
		\caption{Hybrid Inception Net (IN)}
		\label{Fig9F}
	\end{subfigure}
    \begin{subfigure}{0.32\textwidth}
		\includegraphics[width=0.99\textwidth]{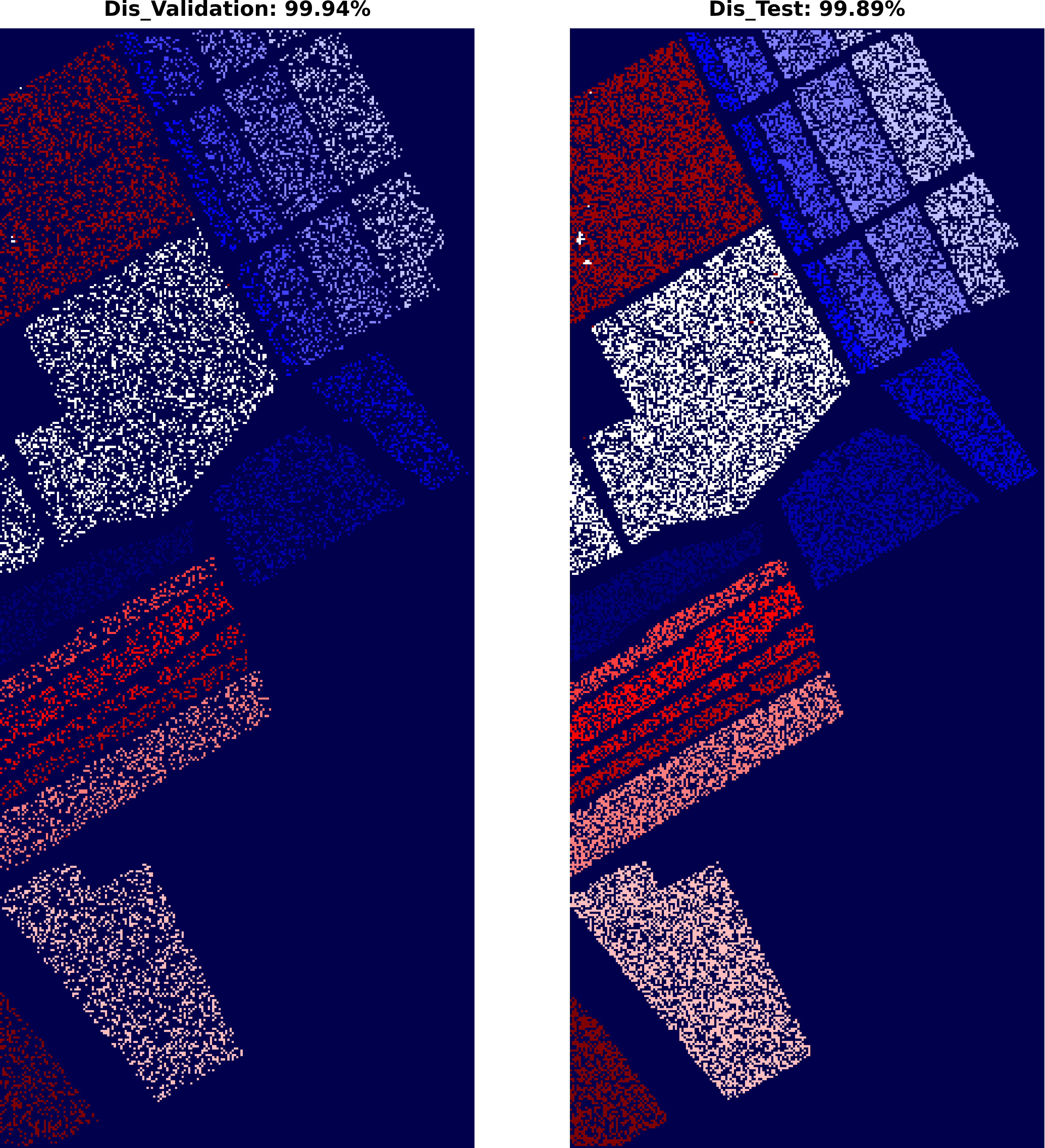}
		\caption{(2+1)D Extreme Exception Net (XN)}
		\label{Fig9G}
	\end{subfigure} 
	\begin{subfigure}{0.32\textwidth}
		\includegraphics[width=0.99\textwidth]{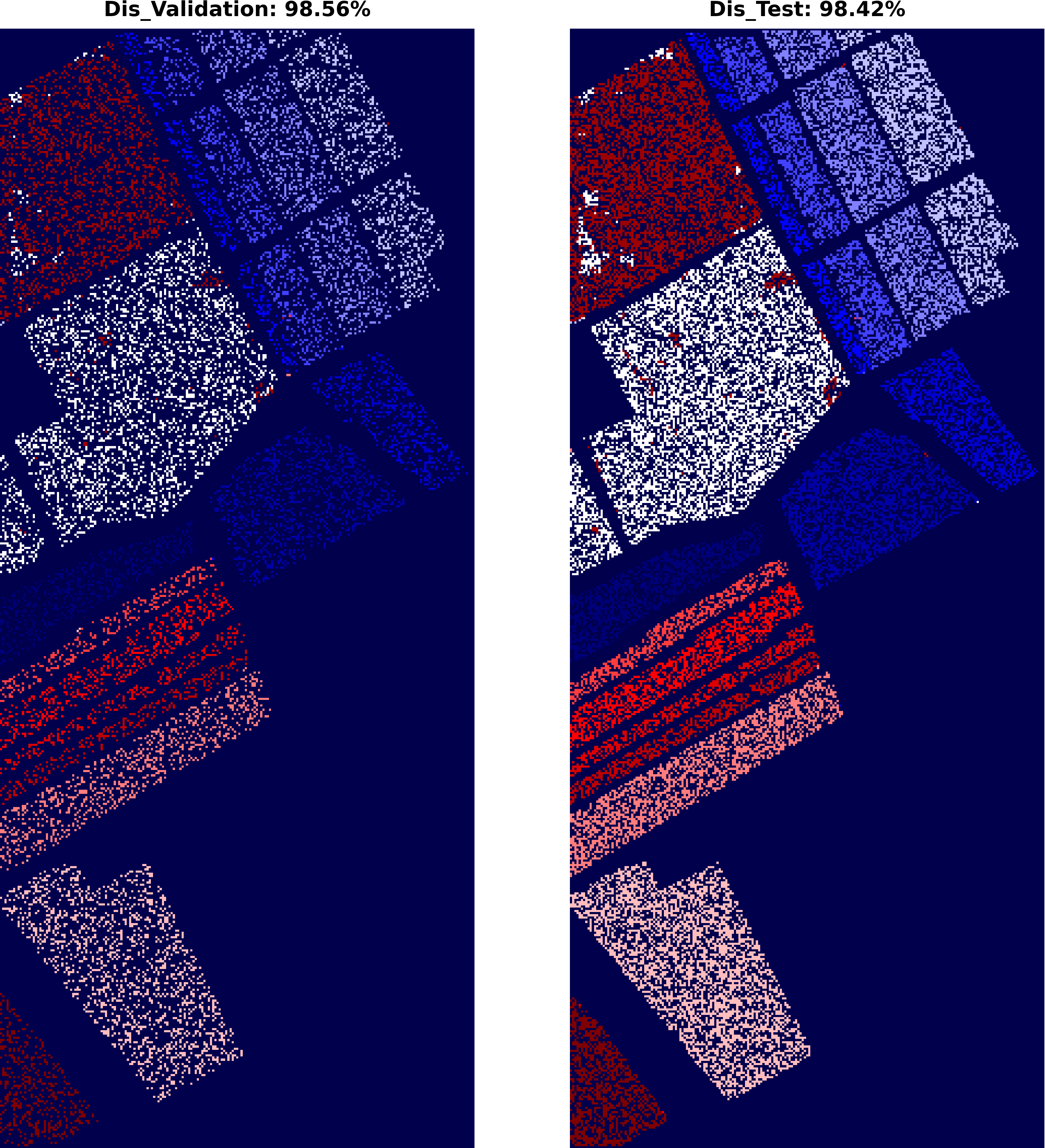}
		\caption{Attention GCN}
		\label{Fig9H}
	\end{subfigure} 
	\begin{subfigure}{0.32\textwidth}
		\includegraphics[width=0.99\textwidth]{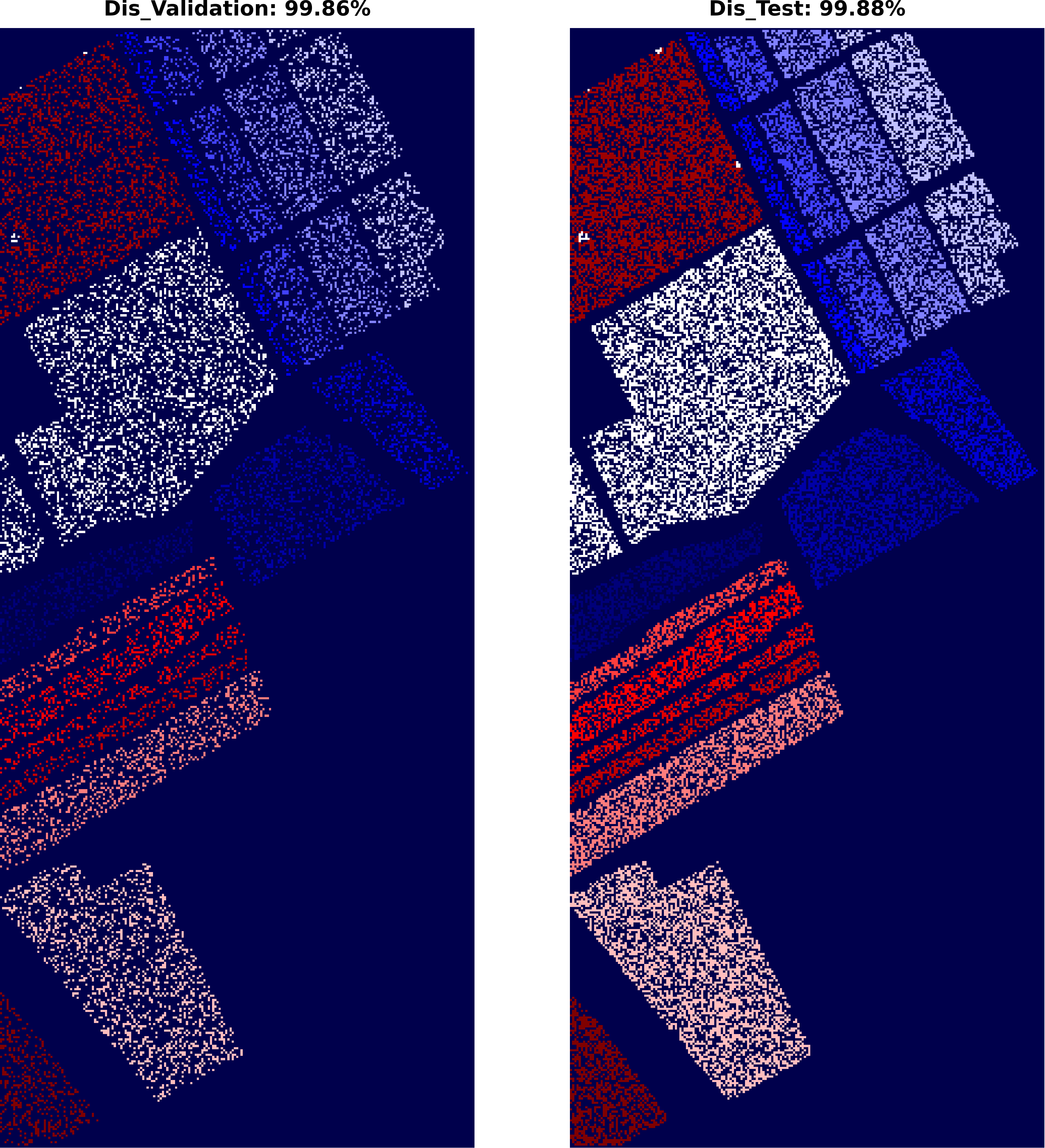}
		\caption{Proposed}
		\label{Fig9I}
	\end{subfigure} 
\caption{\textbf{Salinas Dataset:} Land cover maps for disjoint validation and test set are provided. Comprehensive class-wise results can be found in Table \ref{Tab6}.}
\label{Fig9}
\end{figure*}

The comparative models include; A traditional 2D CNN was used as a baseline model for comparison. This model processes the HSI data as 2D images, ignoring the spectral information. A 3D CNN was employed to capture the spatial-spectral characteristics of HSI data effectively. This model processes the HSI data as 3D volumes, considering both spectral and spatial dimensions. The hybrid CNN combines both 2D and 3D convolutional layers to exploit both spectral and spatial information. The 2D Inception Net architecture, known for its ability to capture multi-scale features, was used for HSIC. The 3D Inception Net architecture extended the 2D Inception Net to capture both spectral and spatial information simultaneously. The hybrid Inception Net combined 2D and 3D Inception modules to leverage both spectral and spatial characteristics. (2+1)D Extreme Expansion Net employed a temporal expansion strategy to handle the temporal dimension of HSI data effectively. An attention-based graph CNN that captures the spatial relationships between pixels was used for HSIC.

The experimental results of attentional feature fusion for HSIC outperformed traditional 2D and 3D CNNs, Hybrid CNNs, 2D and 3D Inception Nets, (2+1)D Extreme Expansion Nets, and Attention Graph CNNs in terms of accuracy, robustness to noise, and interpretability. The attention mechanism in the fusion model effectively captured the spatial-spectral characteristics of HSI data, leading to improved classification performance. The computational efficiency of the fusion model was also competitive, making it a promising approach for HSIC tasks.

Tables \ref{Tab3}, \ref{Tab4}, \ref{Tab5}, and \ref{Tab6} provide a detailed overview of the classification accuracies obtained by the proposed method and several other methods on the IP, SA, UH, and PU datasets. The evaluation metrics encompass OA, AA, $\kappa$ coefficient, and per-class accuracy for each class. Notably, the superior outcomes for each metric are highlighted in bold. The consistent trend observed across all experiments underscores the superior classification performance of the proposed method. Specifically, the proposed method consistently outperforms the comparison methods, securing the highest OA, AA, and $\kappa$ values in each experimental scenario.

The HSI datasets present challenges with their high-resolution images capturing complex urban areas, featuring discrete samples, intricate contextual information, and diverse target scales. Traditional CNN, Inception, and Xception networks face difficulties in effectively processing such intricate data. Although Hybrid Inception Net (Hybrid IN) and (2$+$1)D Extreme Exception Net have shown promising results, Hybrid IN, being CNN-based, encounters challenges in capturing comprehensive global information. On the other hand, Hybrid IN struggles to accommodate the multi-scale nature inherent in HSI data. The attentional features fusion of the proposed pipeline addresses these issues by enabling the model to learn global dependencies across objects at various scales. This capability contributes to the proposed model's superior accuracy compared to a range of convolutional networks.

In the experiments conducted at the PU, the proposed method exhibits a notable advantage over other methods, particularly in classifying targets with spatially similar features like meadows, gravel, bare soil, and bitumen. While 3D CNN, 2D Inception Net, 3D Inception Net, Hybrid Inception Net, and (2+1) Extreme Exception Net show comparable overall classification accuracy, our approach surpasses it in the specific classification of these targets, achieving higher accuracies. This superiority is attributed to the incorporation of fusion within our method, enhancing the effective exploitation of spectral-spatial features.

In the IP experiment, the imbalance between inter-class samples leads to poor performance for 2D CNN, Hybrid CNN, and Attention Graph CNN networks on imbalanced classes. However, the results obtained by the proposed method not only exhibit the highest overall accuracy but also demonstrate a more uniform performance for each class. This suggests that the joint spatial–spectral features and attentional feature fusion mechanism offer advantages over other methods when dealing with imbalanced datasets. Figures \ref{Fig6}, \ref{Fig7}, \ref{Fig8}, and \ref{Fig9} depict the classification results of various methods across HSI datasets. The proposed method consistently achieves results closely aligned with the ground truth map, outperforming other methods. The classification outcomes of conventional CNN and hybrid models reveal notable noise in the classification of large area classes, reflecting the challenges of machine learning algorithms in effectively utilizing spatial information. In contrast, our proposed method successfully addresses the confusion between bare soil and meadows in the PU dataset and reduces "speckles" produced by other methods. Similarly, the classification results for the UH dataset obtained by the proposed method demonstrate superior accuracy compared to other methods.

\section{Conclusions and Future Research Directions}

With the integration of attention mechanisms, the proposed model effectively captures the complex spatial-spectral patterns inherent in HSI data, leading to superior classification performance compared to traditional methods. The experimental results have highlighted several key advantages of the attentional feature fusion model. Firstly, it achieves higher classification accuracy, showcasing its ability to learn discriminative features and extract relevant information from the spectral and spatial dimensions of HSI data. Secondly, it exhibits improved robustness to noise, enabling reliable classification even in the presence of noise and variations within the data. Additionally, the attention mechanism enhances interpretability by highlighting the significant spectral bands and spatial regions contributing to the classification decisions, providing valuable insights for remote sensing applications.

There are several promising directions for future research in HSIC, for instance, optimization which may involve refining the attention mechanisms, exploring different fusion strategies, or investigating novel architectures specifically designed for HSI data. Moreover, the fusion of spatial and spectral attention can be further explored to capture the contextual relationships between neighboring pixels in both spatial and spectral dimensions. This can potentially lead to a better understanding and utilization of the spatial-spectral characteristics of HSI data. Enhancing the interpretability of the attentional feature fusion model is an important research direction. Developing techniques to provide more detailed and understandable explanations for classification decisions will enhance the trust and usability of the model in real-world applications. In a nutshell, future research should focus on optimizing the model, exploring transfer learning and domain adaptation, enhancing explainability, addressing data scarcity through data augmentation, incorporating spatial-spectral context, and applying the model to real-world scenarios. These efforts will contribute to further advancements in HSIC and facilitate its practical implementation in diverse domains.

\bibliographystyle{IEEEtran}
\bibliography{IEEEabrv,Sam}
\end{document}